%% file: main.tex
\documentclass[conference]{IEEEtran}
\IEEEoverridecommandlockouts
\usepackage{cite}
\usepackage{amsmath,amssymb,amsfonts}
\usepackage{algorithmic}
\usepackage{graphicx}
\usepackage{textcomp}
\usepackage[table]{xcolor}
\usepackage{comment}
\usepackage{subfigure}
\usepackage{array}
\usepackage{gnuplottex}
\usepackage{balance}

\usepackage[bookmarks=false]{hyperref} % Removed for IEEE compliance

\def\BibTeX{{\rm B\kern-.05em{\sc i\kern-.025em b}\kern-.08em
    T\kern-.1667em\lower.7ex\hbox{E}\kern-.125emX}}

\makeatletter
\newcommand{\thickhline}{%
    \noalign {\ifnum 0=`}\fi \hrule height 1pt
    \futurelet \reserved@a \@xhline
}
\newcolumntype{"}{@{\hskip\tabcolsep\vrule width 1pt\hskip\tabcolsep}}
\makeatother
\definecolor{light-gray}{gray}{0.95}

\begin{document}

% limit 10 pages excluding references (unlimited for ref).

%
% TODOs: indicate the number of channels for each layer
% 
%

%\title{Optimal Convolutional Neural Network Size for\\Best Performance on Mobile and Embedded GPUs}

%\title{Optimal Convolutional Dimensionality for Mobile and Embedded GPU Performance}

%\title{DNN Library-aware Pruning for Embedded GPUs}

\title{Performance Aware Convolutional Neural Network Channel Pruning for Embedded GPUs}

%\title{On the Pitfalls of Uninstructed CNN Channel Pruning on Embedded GPUs}

\begin{comment}
\author{\IEEEauthorblockN{Valentin Radu}
\IEEEauthorblockA{\textit{University of Edinburgh}\\
valentin.radu@ed.ac.uk}
\and
\IEEEauthorblockN{Kuba Kaszyk}
\IEEEauthorblockA{\textit{University of Edinburgh}\\
Kuba.Kaszyk@ed.ac.uk}
\and
\IEEEauthorblockN{Yuan Wen}
\IEEEauthorblockA{\textit{Trinity College Dublin} \\
weny@scss.tcd.ie}
\and
\IEEEauthorblockN{Jack Turner}
\IEEEauthorblockA{\textit{University of Edinburgh}\\
jack.turner@ed.ac.uk}
\and
\IEEEauthorblockN{Michael O'Boyle}
\IEEEauthorblockA{\textit{University of Edinburgh}\\
mob@inf.ed.ac.uk}
}
\end{comment}

\author{\IEEEauthorblockN{Valentin Radu\IEEEauthorrefmark{1},
Kuba Kaszyk\IEEEauthorrefmark{1},
Yuan Wen\IEEEauthorrefmark{2},
Jack Turner\IEEEauthorrefmark{1},
Jos\'e Cano\IEEEauthorrefmark{3},
Elliot J. Crowley\IEEEauthorrefmark{1}, \\
Bj{\"o}rn Franke\IEEEauthorrefmark{1}, %\\
Amos Storkey\IEEEauthorrefmark{1},
Michael O'Boyle\IEEEauthorrefmark{1}}
\IEEEauthorblockA{
        \begin{tabular}{cc}
                \IEEEauthorrefmark{1}{University of Edinburgh, UK}
                \IEEEauthorrefmark{2}{Trinity College Dublin, Ireland}
                \IEEEauthorrefmark{3}{University of Glasgow, UK}
        \end{tabular}
  }		
}

\maketitle

\input{abstract}

\begin{IEEEkeywords}
convolutional neural networks, channel pruning, embedded GPU
\end{IEEEkeywords}

\input{introduction}
\input{motivation}
\input{setup}
\input{experiments}

\input{discussion}
\input{related_work}

\input{conclusions}

\section*{Acknowledgment}

This project has received funding from the European Union’s Horizon 2020 research and innovation programme under grant agreement No. 732204 (Bonseyes). This work is supported by the Swiss State Secretariat for Education, Research and Innovation (SERI) under contract number 16.0159. The opinions expressed and arguments employed herein do not necessarily reflect the official views of these funding bodies.

\balance
\bibliographystyle{unsrt}
\bibliography{main}

\end{document}

%% file: abstract.tex
\begin{abstract}
Convolutional Neural Networks (CNN) are becoming a common presence in many applications and services, due to their superior recognition accuracy. They are increasingly being used on mobile devices, many times just by porting large models designed for server space, although several model compression techniques have been considered. One model compression technique intended to reduce computations is channel pruning. Mobile and embedded systems now have GPUs which are ideal for the parallel computations of neural networks and for their lower energy cost per operation. Specialized libraries perform these neural network computations through highly optimized routines. As we find in our experiments, these libraries are optimized for the most common network shapes, making uninstructed channel pruning inefficient. We evaluate higher level libraries, which analyze the input characteristics of a convolutional layer, based on which they produce optimized OpenCL (Arm Compute Library and TVM) and CUDA (cuDNN) code. However, in reality, these characteristics and subsequent choices intended for optimization can have the opposite effect. We show that a reduction in the number of convolutional channels, pruning 12\% of the initial size, is in some cases detrimental to performance, leading to 2$\times$ slowdown. On the other hand, we also find examples where performance-aware pruning achieves the intended results, with performance speedups of 3$\times$ with cuDNN and above 10$\times$ with Arm Compute Library and TVM. Our findings expose the need for hardware-instructed neural network pruning.
\end{abstract}

% Convolutional Neural Networks (CNN) are becoming a common presence in many applications and services, due to their superior recognition accuracy. They are increasingly being used on mobile devices, many times just by porting large models designed for server space, although several model compression techniques have been considered. One model compression technique intended to reduce computations is channel pruning. Mobile and embedded systems now have GPUs which are ideal for the parallel computations of neural networks and for their lower energy cost per operation. Specialized libraries perform these computations to accelerate neural network execution through highly optimized routines. As we find in our experiments, these libraries are optimized for the most common network shapes, making uninstructed channel pruning inefficient. We evaluate higher level libraries, which analyze the input characteristics of a convolutional layer, based on which they produce optimized OpenCL (Arm Compute Library and TVM) and CUDA (cuDNN) code. However, in reality, these characteristics and subsequent choices intended for optimization can have the opposite effect. We show that a reduction in the number of convolutional channels, pruning 12\% of the initial size, is in some cases detrimental to performance, leading to 2$\times$ slowdown. Otherwise, when pruning is performed with a view of the library performance speedups of 3$\times$ with cuDNN and above 10$\times$ with Arm Compute Library and TVM are achievable. Our findings expose the need for hardware-instructed neural network pruning.

%% file: introduction.tex
\section{Introduction}\label{introduction}
%{\color{red}[Budget 1.0 pages.]}

Due to their superior recognition accuracy, Convolutional Neural Networks (CNN) are dominant in several disciplines: computer vision (for image classification~\cite{alexnet,He_2016_CVPR,denseNet2017}, image segmentation~\cite{7298965,10.1007/978-3-319-24574-4_28}, objects in image detection~\cite{7485869,yolo2016}, image style transfer \cite{7780634}, etc.), speech recognition~\cite{Zhang2016TowardsES} and natural language processing~\cite{cnnsentence2014,Kalchbrenner14aconvolutional}.

These solutions are making their way into smaller devices, on mobile phones and home personal assistant devices. %~\cite{google_news}.
However, current CNN models are still too large for immediate deployment on resource-constrained devices. Pruning is a widely accepted practice to make these large models suitable to run on such small devices. It is well understood in the machine learning community that neural networks can produce good inferences even after pruning a substantial amount of their internal parameters (weights)~\cite{NIPS1988_156,han2015deep,han2015learning}. In \textit{Channel Pruning}, entire channels (or filters) are assessed for their importance to determine if these may be removed~\cite{he2017channel} to produce a slimmer network from the original one, with minimal drop in inference accuracy. Unlike other pruning methods, this produces a compact dense network suitable for the already optimized dense convolutional routines~\cite{our18iiswc}.

Currently, only accuracy is considered in the iterative loop of channel pruning, removing channels and retraining to compensate for loss. This process is agnostic to target devices, expecting that having a smaller number of network parameters will lead to faster inference at deployment. Contrary to this expectation, we find that uninstructed channel pruning can hurt performance dramatically, up to 2$\times$ slowdown in some cases when pruning just 12\% of layer channels. We develop the case that inference time on the target device should also be considered when producing smaller networks through channel pruning. %This is cause by library optimisations that consider only some aare just a on running a pruned nethis is actually important to know and is very complex to determine offline. Execution profiles done on larger GPUs are not representative of the conditions on deployment embedded GPUs. Deep learning libraries and hardware specifications affect the execution times, so these should also be considered when selecting the best pruning level.

The parallel nature of computations required by neural networks exposes GPUs as the compute unit of choice, including on mobile and embedded systems for superior FLOPS per watt performance. Dominant in this space are Arm Mali GPUs and Nvidia embedded Jetson GPUs, each programmed via different computing libraries (OpenCL and CUDA). %for optimal performance. 
These are called by higher level libraries, such as Arm Compute Library (ACL) and cuDNN. However, not much is known about the performance of these libraries on custom deep learning workloads.

Here we expose the characteristics of higher level libraries used for deep neural network computations on embedded GPUs, showing their unintuitive behavior in response to changes to convolutional layer size. We experiment with three deep learning libraries, Arm Compute Library and TVM for Mali GPUs and cuDNN for Jetson embedded GPUs, on four different devices, observing unintuitive performance patterns caused by their internal heuristics. Intrigued by these observations, we take an in-depth perspective by highlighting these patterns on a Mali GPU simulator where we find that bad splits of convolutional workload into multiple kernels adds substantial overhead, hurting performance. %We find that how these libraries choose to divide the workload into GPU kernels has a big impact on performance on these devices.

Our findings are relevant in both the systems and machine learning communities. First, it is important to understand the impact of pruning on inference time, not just classification accuracy, and to identify how the number of channels can be calibrated to improve on both metrics simultaneously. Second, designing new neural network architectures for specific devices should consider the best sizes of convolutional layers for each library and hardware, thus building specialized networks for each runtime environment. And third, library heuristics for workload optimization should be revisited to capture the increasing variation of neural networks and computing devices.

In this paper we make the following contributions:

\begin{itemize}
    \item We expose the behavior of three popular deep learning libraries on varying convolutional layer sizes across four different devices.
    \item These run-time performances are analyzed in-depth through a GPU simulator to understand the built-in heuristics for optimizations and how this performs unjustified splits of workload hurting performance.
    \item On the observed staircase-like performance pattern, we propose the selection of optimal convolutional size in an iterative loop with hardware profiling and test accuracy of the compressed model.
    %\item With these empirical observations we demonstrate that best inference time is achievable by selecting between library implementations for each layer of a neural network since no library outperforms at across all layers on embedded GPUs.
\end{itemize}

The remainder of this paper is organized as follows. Section~\ref{sec:motivation} offers background into convolutional neural networks and channel pruning, and motives our work on these. Section~\ref{sec:setup} provides the experimental setup. Section~\ref{sec:experiments} presents our ample experiments, followed by a discussion on observations from these in Section~\ref{sec:discussion}. Finally, we present related work in Section~\ref{sec:related_work} and conclude in Section~\ref{sec:conclusions}.

%%%%%%%%%%%%%%%%%%%

\begin{comment}
the which On these new devices these models are run with native computation libraries for efficient neural network computation (Arm Compute Library,  from their unconstrained computation resources to mobile devices where these run with local computation libraries available on these smaller devices. Although much consideration has been given to designing these libraries to be efficient on these resource restricted devices, we find that often these are tuned to handle the more popular cases of deep neural networks. The common approach to make CNNs available on smaller devices is to compress their structure by channel pruning. However, we find that that Through broad experimentation we discover that these are not optimal for all cases. In this work we evaluate the performance of popular DNN libraries for embedded GPUs -- Arm Compute Library, TVM and cuDNN. We find that some steps on the pruning staircase of convolutional layers are more efficient than others, so optimisation to pick these points are desirable. In some cases for similar number of parameters the performance can be improved by almost 3$\times$ with common DNN libraries, which should be strongly considered when designing and pruning convolutional neural networks for embedded devices.
\end{comment}

%% file: motivation.tex
\section{Background and Motivation}\label{sec:motivation}
%{\color{red}[Budget 1.5 pages.]}

\subsection{Background}

\subsubsection{Convolutional Neural Networks}

Convolutional Neural Networks have been adopted for most computer vision tasks. These are composed of stacked convolutional layers with multiple filters (or channels) which are convolved over an input image to produce a multi-channel output (each filter producing an output channel). Besides convolutional layers, CNNs can encompass other layer types to produce different affine transformations on their input tensor (such as Dropout, activation layers -- ReLU, Tanh, etc.). Although important, these affine transformations
account for very little in the total inference time of modern neural networks, with most of the computational load being executed in the convolutional layer. 
The recent ImageNet winner, SENet~\cite{hu2018squeeze} is predominantly formed of convolutional layers, these accounting for 99.991\% of total floating point operations. For this reason convolutional layers have received much attention, with different optimizations being proposed in deep learning libraries to accelerate their execution.

Several routines exist to perform the convolution operation, although two are dominant across the majority of libraries:
\begin{itemize}
    \item Direct convolution -- this method shifts each filter (channel) one position at a time over an input image with a deep nested loop. This requires the least amount of extra memory, which makes it ideal for devices with limited physical memory, although it is also very slow in terms of computation time.
    \item General Matrix Multiplication (GEMM) -- this method performs the convolution by unrolling each image patch to convolve over into a column of a larger matrix of unrolled patches, while filters (channels) are unrolled into rows to form a second large matrix, in a process known as \textit{image2col}~\cite{jia2014learning}. The entire convolutional operation over the input image is performed by a single operation of matrix to matrix multiplication on the two large matrices resulting from the unrolling process mentioned earlier. This is a very popular approach due to the readily available, highly optimised matrix multiplication libraries (Blas, CUDA), which make it fast in practice.
\end{itemize}

\subsubsection{Channel Pruning}

\begin{figure*}
    \centering
    \includegraphics[width=0.85\textwidth]{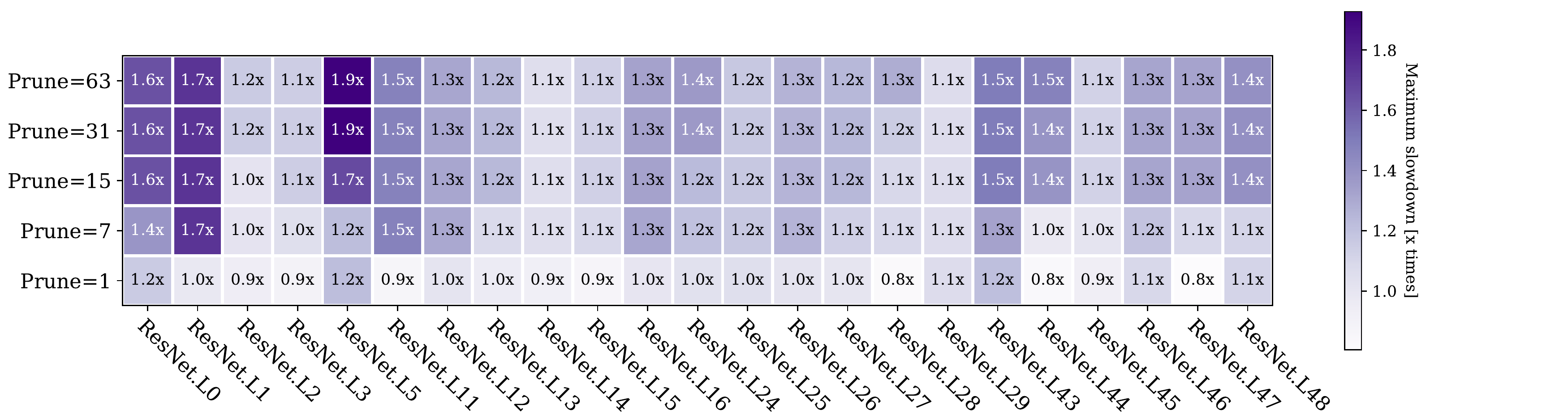}
    \caption{Potential slowdown in execution time of pruned network layers compared to original large model when pruning a number of channels \textit{(Prune)} from the initial number of channels for each convolutional layer of ResNet-50. Performance observed when running on a mobile GPU (Mali G72).}
    \label{fig:heat_slowdown}
\end{figure*}

\begin{figure}
    \centering
    \includegraphics[width=0.9\columnwidth]{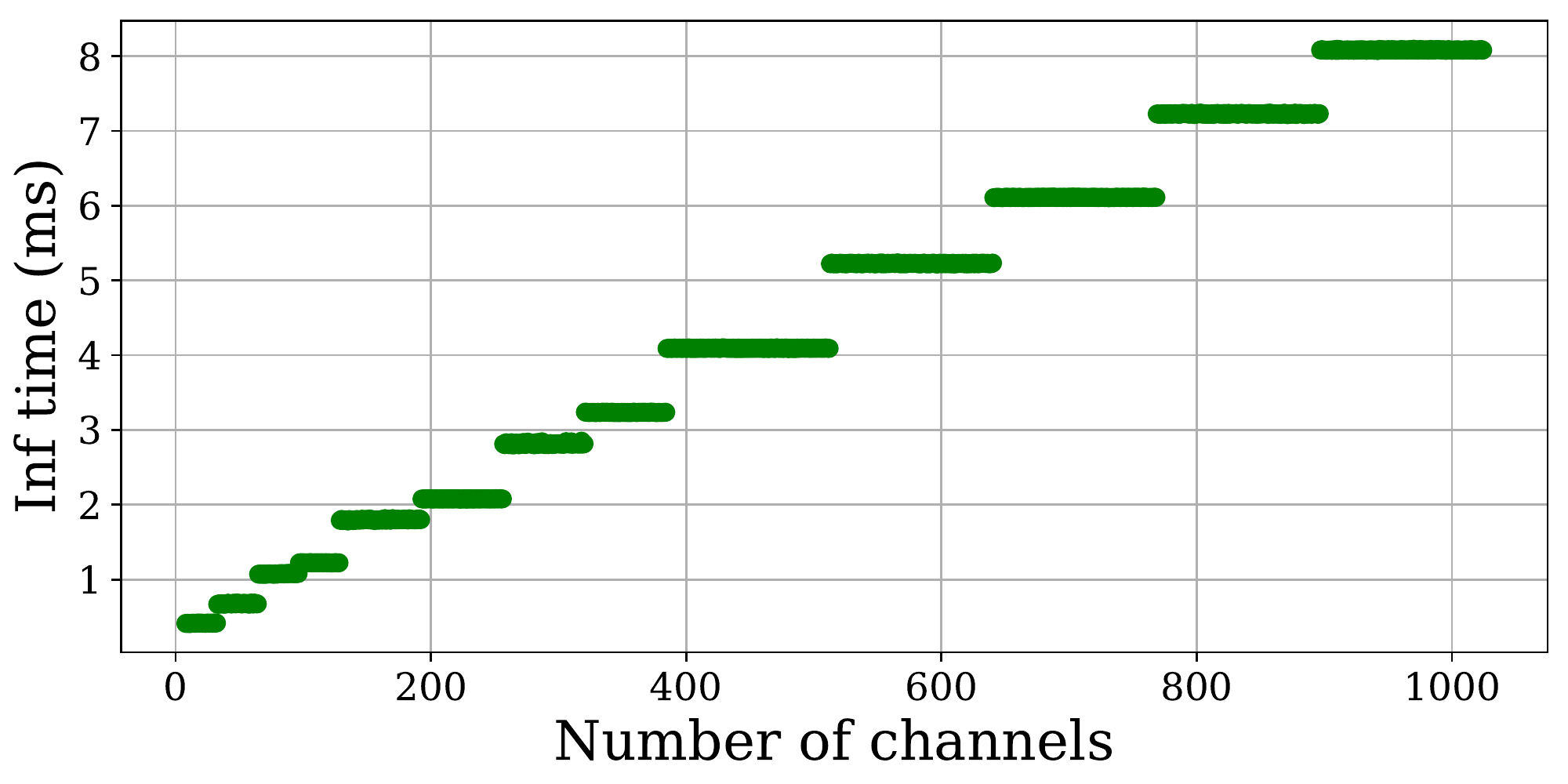}
    \caption{Staircase correlation between inference time and number of parameters (channels) in a ResNet-50 layer on the Jetson TX2, showing a more intuitive performance pattern.}
    \label{fig:nvidia_staircase}
\end{figure}

\begin{figure}
    \centering
    \includegraphics[width=0.9\columnwidth]{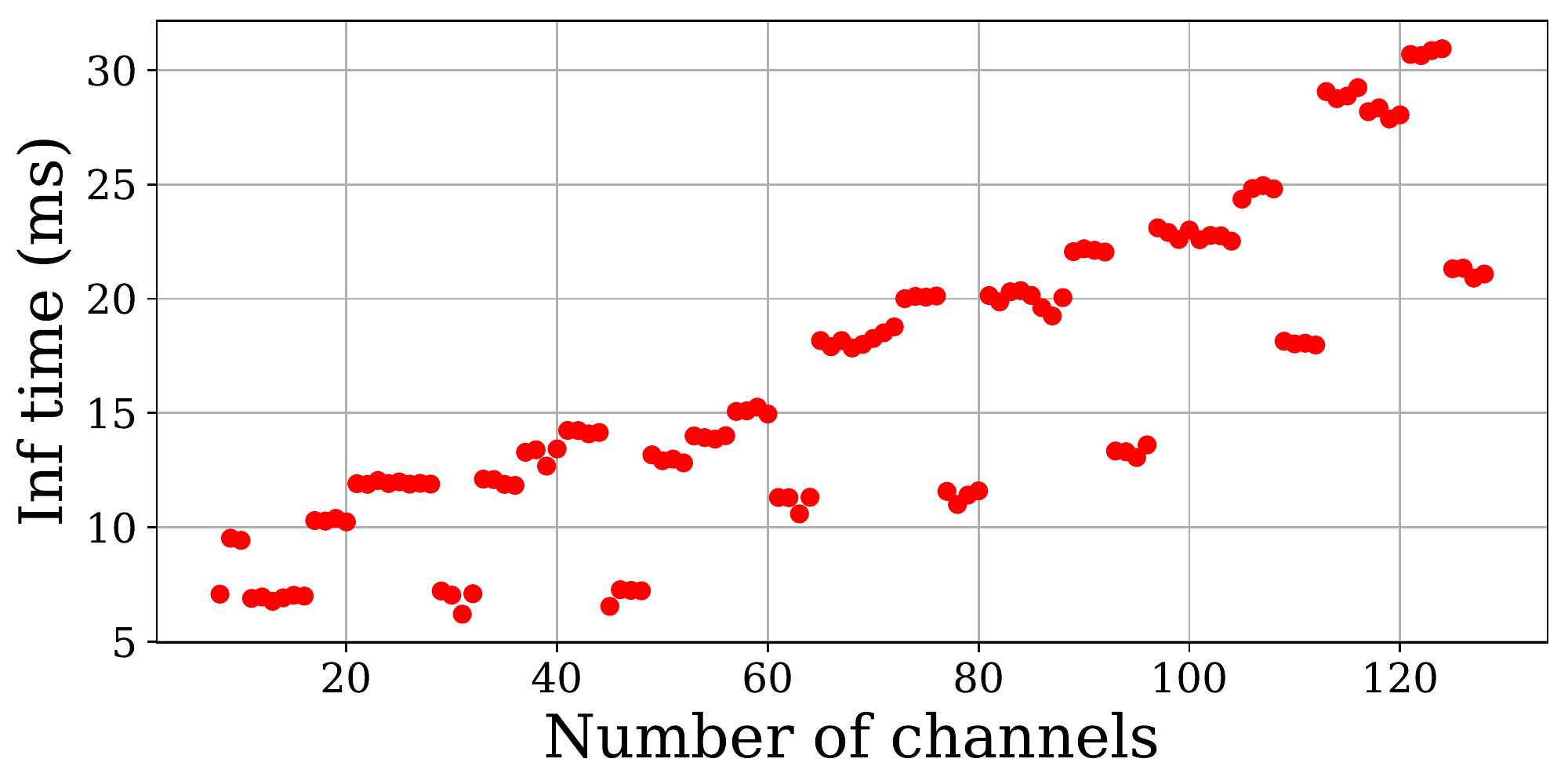}
    \caption{Inference time of a convolutional layer of ResNet-50 run with the Arm Compute Library with varying amount of channel pruning.}
    \label{fig:hikey_staircase}
\end{figure}

Current large CNNs require some alteration to make them suitable for deployment on smaller devices, which often comes in the form of pruning. Weight pruning, through which some weights based on a signal are reduced to zero~\cite{han2015deep}, is one approach that works well with accelerators of sparse algebraic operations, although the speedup these can offer on general purpose devices has been questioned~\cite{our18iiswc}. Another approach for network size reduction is \textit{channel pruning}, in which entire channels are eliminated if their impact is minimal~\cite{he2017channel}, resulting in better performance than other compression techniques~\cite{our18iiswc}, and can be modeled with both accuracy and inference time constraints~\cite{our18hakd}.

\subsection{Channel Pruning on Different GPUs}

As a machine learning technique, CNN pruning is generally performed away from the runtime environment, with the primary metric for the task being inference accuracy. Retraining the model during the pruning process requires substantially more computing resources so this is generally performed on other machines than the final inference device.

In this work we perform channel pruning without considering the accuracy impact, but our channel pruning approach has the same effect on inference time as when done with accuracy conditions. Assuming the $c$-th convolutional layer of a neural network has $n$ filters (channels) $k_{i}$, $i\in[1,n]$ (before pruning). To prune channel $p$, with $1\leq p \leq n$, the new convolutional layer will have a number of $n-1$ channels and each channel $k_{i}, i\in[p+1,n]$ will be re-indexed to $i=i-1$. For example, in a convolutional layer with 128 channels, pruning the 25-th channel will produce a compact layer with channel 26 becoming channel 25, and so on for the following channels re-indexing to $i-1$, thus producing a new convolutional layer with channels indexed continuously from 1 to 127. This process is repeated for each pruned channel. As can be observed, by this process the same computation time will be produced no matter which channel is picked for pruning, so we eliminate channels sequentially for our inference time analysis.

By observing the execution time of different pruning levels of a ResNet-50 convolutional layer (presented in Figure~\ref{fig:nvidia_staircase}) on a Jetson TX2, a staircase shape pattern emerges. There are stepped changes in inference time by varying the number of channels, due to filling the workgroup sizes on the device. These gaps can lead to a substantial penalty in execution time between layer configurations with similar numbers of channels. Ideally, one should aim to choose the number of channels of a convolutional layer such that it falls to the right side of a performance step (more channels for the same execution time budget), as we explored in another work based on inference accuracy~\cite{our18hakd}. A larger number of channels holds more parameters in a neural network, usually leading to higher accuracy in the prediction task, as also reflected by the current trend in machine learning to design increasingly larger networks for better accuracy.

However, these well-shaped patterns are not representative for all mobile and embedded GPUs. For instance, the execution of pruned layers of ResNet-50 on a Mali G72 GPU implemented with the Arm Compute Library (Figure~\ref{fig:hikey_staircase}) shows a pattern with two parallel staircases. This can have severe consequences depending on which performance step the pruned layer falls on. In fact, pruning risks introducing slowdown in execution time, with pruned networks potentially running slower than the original unpruned larger network, if libraries and hardware performance are not considered in the pruning process. This situation is presented in Figure~\ref{fig:heat_slowdown} for running an implementation of pruning with the Arm Compute Library using the GEMM method on the HiKey 970. Pruning at a distance of only 64 channels can match a performance step that introduces up to 2$\times$ slowdown in execution time compared to the initial layer (unpruned). Intuitively, some performance steps will offer speedups, but having some levels of pruning that can lead to slowdowns is hazardous and contrary to our expectation that using pruning (fewer network parameters and operations) will produce an universally faster network for any device and with any deep learning libraries.

This unintuitive behavior of deep learning computing libraries, each driven by their own internal optimisations is what motivates this exploration. In the following sections we expose the optimal number of channels for a few deep neural networks, with a range of deep learning libraries and on various devices, expressing the speed-ups achievable by performance aware pruning. 

% By contrast, when running the convolutional layer with the Arm Computer Library on a Mali G72 GPU, we observe a different trend for the execution time of varying pruning. Figure~\ref{fig:hikey_staircase} presents the execution time for different layer size after pruning, and we can observe two parallel staircases, in groups of four on the faster (in inference time) staircase and three consecutive groups of fours on the slower staircase. These are specific to the library implementation and how these optimize for workgroup sizes and memory hierarchy.

% We consider matching the library and hardware conditions critically important for mobile and embedded devices with limited resources, some being battery powered. While the research focus has been on striking the balance between number of parameters and model accuracy, here we expose the importance of considering the execution environment in addition to this balance, so that it can inspire which are the optimal pruning levels.

%% file: setup.tex
\section{Experimental Setup}\label{sec:setup}

\subsection{Neural Network Libraries}
%{\color{red}[Budget 1.25 pages.]}

We explore the most common libraries used in programming embedded GPUs for neural network workloads, focusing on two libraries that generate OpenCL code (Arm Compute Library and TVM) to run on Mali GPUs and a CUDA library (cuDNN) for Nvidia embedded GPUs.

\subsubsection{Arm Compute Library (ACL)} Is a collection of functions and APIs to program Arm CPUs and Arm Mali GPUs through OpenCL, applying low-level optimizations for best performance on Arm mobile and embedded processors. We used version v19.02 in our experiments.

\subsubsection{TVM}
Also an OpenCL based framework is TVM. This is an open source deep learning compiler stack performing several optimisations at each level in the stack, including compute graph optimization for operator fusion, layout transformations, and memory management. We used version 0.6 here.

\subsubsection{cuDNN}
Is an Nvidia proprietary library (v7) containing efficient implementations of CUDA primitives to run deep learning workloads. It is used for programming both embedded GPUs (Jetson) and desktop Nvidia GPUs.

\subsection{Models}

To generalize the observation of pruning patterns we select three popular deep neural networks prevalent in computer vision for image classification:
\begin{itemize}
    \item ResNet-50~\cite{he2016deep} has 50 layers and consists of residual blocks. There are 23 convolutional layers with filters of size $3\times3$ and $1\times1$ (referred to as ResNet.L$i$, where $i$ is the layer index), and interleaved with other layers, such as batch normalization. Although they are indexed, we do not profile their performance here due to their cost being insignificant. Convolutional layers have a number of filters between 64 and 2048~\cite{he2016deep}.
    \item VGG-16~\cite{vgg}, is a feed-forward network with 13 convolutional layers and 3 fully connected layers. Each convolution uses $3\times3$ size filters. The convolutional layers are indexed similarly to ResNet, with {0, 2, 5, 7, 10, 12, 17, 19, 24} unique shapes (where the convolutional layer shape is repeated in the network, it is considered only once). These convolutional layer have the following number of filters: 64, 64, 128, 128, 256, 256, 512, 512, and 512 respectively.
    \item AlexNet~\cite{alexnet} is the earliest CNN to win the ImageNet competition by a huge margin over the previous top machine learning solution. Compared to more recent CNNs this has only 5 convolutional layers, indexed 0, 3, 6, 8, 10 interleaved by Pooling and Dropout layers. The unpruned convolutional layers have the following number of filters: 64, 192, 384, 256, and 256 respectively.
\end{itemize}

\subsection{Profilers}

\subsubsection{OpenCL} We developed a custom library to intercept each OpenCL call in order to observe the time when OpenCL kernels start their execution on the GPU and when this finishes, for a precise assessment of their runtime. We can also inspect the name of each kernel and memory footprint to ensure we measure the correct kernel.

\subsubsection{CUDA} We measure the time between CUDA events to determine the execution time for each cuDNN task. These times were compared and matching those reported by the official Nvidia \textit{nvprof} profiler.

\subsection{Devices}

% \begin{figure*}[ht] %
%   \centering
%   \subfigure[HiKey 970]{%
%   \includegraphics[width=0.2\textwidth]{images/hikey.jpg}
%   \label{fig:hikey}%
%   }\hfil
%   \subfigure[Odroid XU4]{%
%   \includegraphics[width=0.2\textwidth]{images/odroid.jpg}
%   \label{fig:odroid}%
%   }\hfil
%   \subfigure[Nvidia Nano]{%
%   \includegraphics[width=0.2\textwidth]{images/nvidia_nano.jpg}
%   \label{fig:nano}%
%   }\hfil
%   \subfigure[Nvidia Jetson]{%
%   \includegraphics[width=0.2\textwidth]{images/jetson.jpg}
%   \label{fig:jetson}%
%   }
%   \caption{Experiments conducted on development boards with embedded GPUs, using OpenCL to programme the first two and CUDA for the Nvidia boards.}\label{fig:devices}
% \end{figure*}

We experiment on four different devices: two with Mali GPUs (HiKey 970 with the Mali G72 architecture, and Odroid XU4 with the Mali T628 architecture) programmed with OpenCL through the ACL; and two with the Nvidia Jetson embedded GPU (TX2 and Nano), all with default OS. The median time of 10 runs is reported for each configuration.

%% file: experiments.tex
\section{Experiments}\label{sec:experiments}
%{\color{red}[Budget 5 pages.]}

\begin{figure}
    \centering
    \includegraphics[width=0.9\columnwidth]{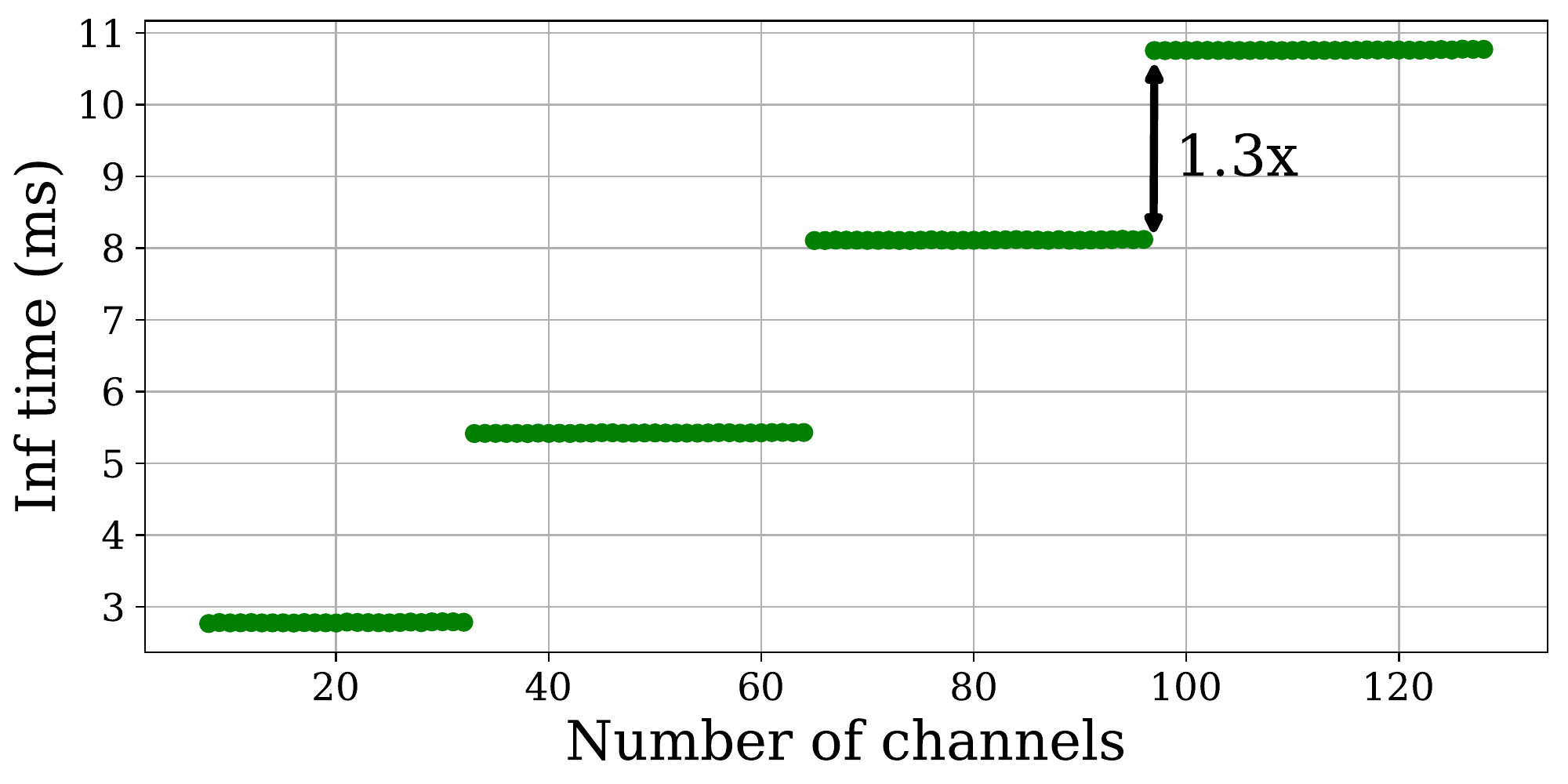}
    \caption{Staircase (execution pattern) observed for channel pruning layer 16 of ResNet-50 implemented with CuDNN on \textit{Jetson TX2}.}
    \label{fig:tx2_layer16}
\end{figure}

\begin{figure}
    \centering
    \includegraphics[width=0.9\columnwidth]{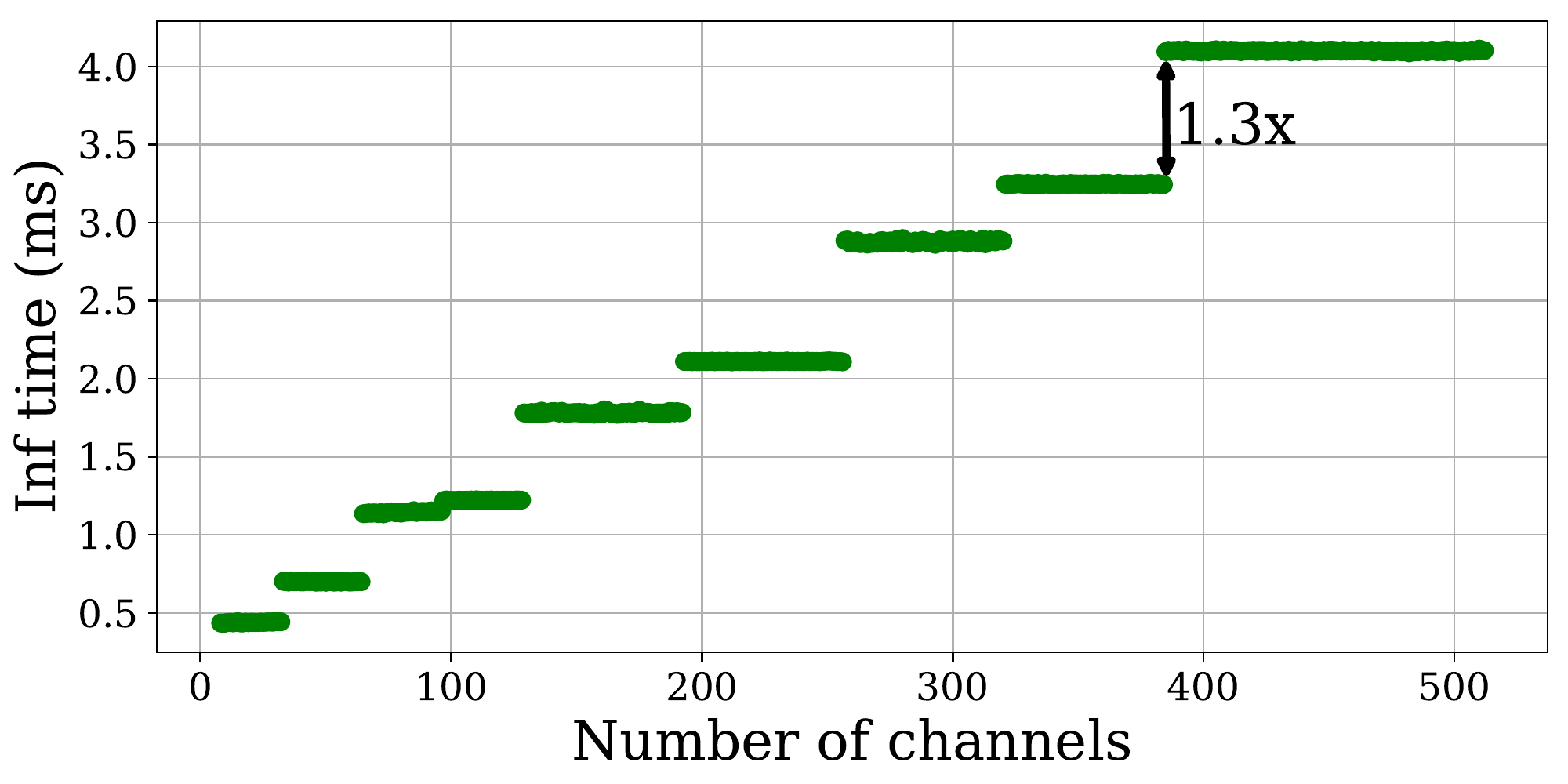}
    \caption{Staircase (execution pattern) observed for channel pruning layer 14 of ResNet-50 implemented with CuDNN on \textit{Jetson TX2}.}
    \label{fig:tx2_layer14}
\end{figure}

\begin{figure*}
    \centering
    \includegraphics[width=0.85\textwidth]{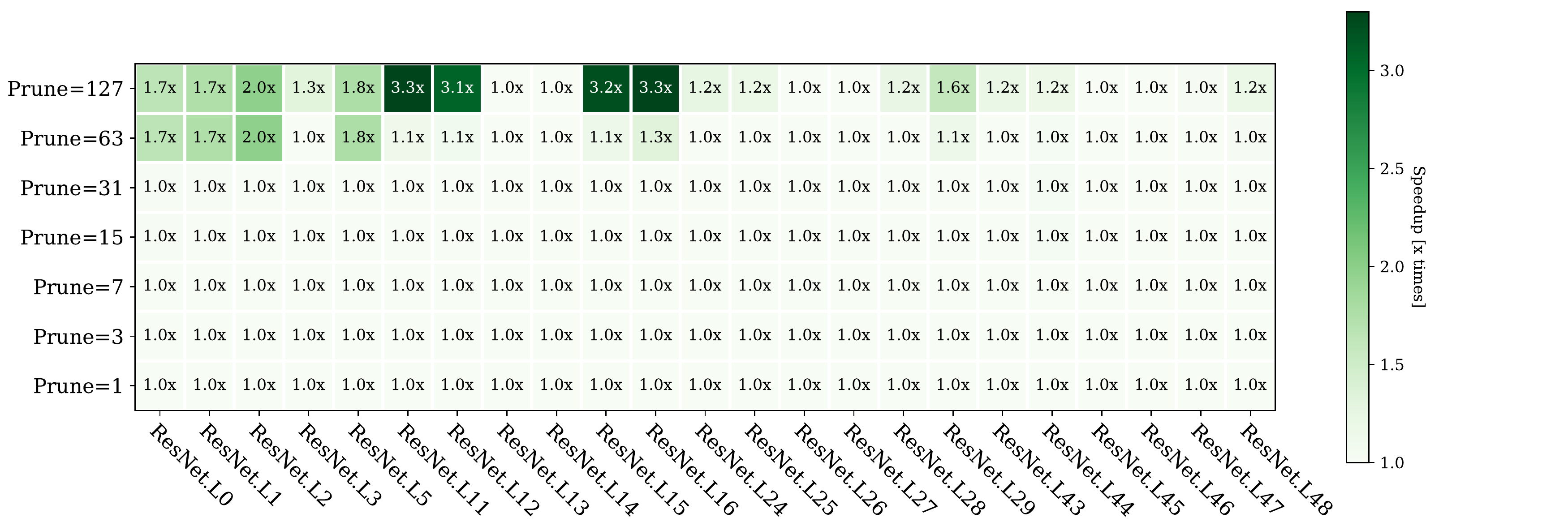}
    \caption{Speedups observed when pruning at different distances within each layer of ResNet-50 using the CuDNN implementation running on Jetson TX2.}
    \label{fig:cudnn_resnet_heatmap}
\end{figure*}

This section presents the profiled performance of pruned CNN layers on embedded GPUs, and an in-depth analysis of the ACL through a GPU simulator to explain the observed behavior. We first profile these layers with cuDNN on the Jetson platforms, followed by ACL with both Direct Convolution and GEMM method on devices and in the GPU simulator, and finally using TVM optimized code on Mali GPUs. %To the end of this section we have an analysis of the best network configuration selecting implementation for each layer from across libraries.

% [..TODOs:]

% - description of the tool used to intercept OpenCL calls

% - Why ResNet? present the structure of each ResNet layer.

% This is a representative workload since it is used for many tasks beyond image classification, it is formed of 3x3 convolutions, which are common to many other networks (VGG, MobileNet, etc.) and is the inspiration for new architectures DenseNet, WideResNet, etc.

% The structure of convolutional layers is presented in Table~\ref{tab:resnet_structure}.

% \begin{table}[h]
% \centering
% \small
% \rowcolors{2}{white}{light-gray}
% \begin{tabular}{lrr}
% \thickhline
% \bf{Layer id} & \bf{input size (CxHxW)} & \bf{C}\\ \hline
 
% 1 & 1,900,416 & 169,344\\ \thickhline
% \end{tabular}
% \caption{ResNet-50 convolutional layers structure.}
% \label{tab:resnet_structure}
% \end{table}

\subsection{Convolutional Layer Pruning Staircase}

We explore the runtime performance of channel pruning for all convolutional layers of three popular CNNs (ResNet-50, VGG-16 and AlexNet) by gradually reducing the number of channels of each layer, one at a time, and observing the performance of the new layer on devices. We present the performance of each deep learning library separately.

\subsubsection{CuDNN on Mobile GPUs}

We profile gradual channel pruning for each layer of ResNet-50. Figure~\ref{fig:tx2_layer16} presents the pattern in pruning layer 16 of ResNet-50, showing a flat performance (same inference time) for all the channels above 97, with a drop in inference time for a layer of 96 channels (and fewer), with a speedup of 1.3$\times$. Another such drop in inference time is for a layer with 64 channels and below. This layer has four optimal execution points, to the right of each stair (most number of channels for an inference time), which should be considered when pruning to offer the best trade-off between accuracy and inference time.

A similar behavior is observed for channel pruning on layer 14, also from ResNet-50, this time with more stairs due to the larger number of channels this layer starts with, but also with a different drop in inference time between these steps (Figure~\ref{fig:tx2_layer14}). This uneven gap in inference time between stairs should also be considered when assessing the optimal level of channel pruning.

\begin{figure}
    \centering
    \includegraphics[width=0.9\columnwidth]{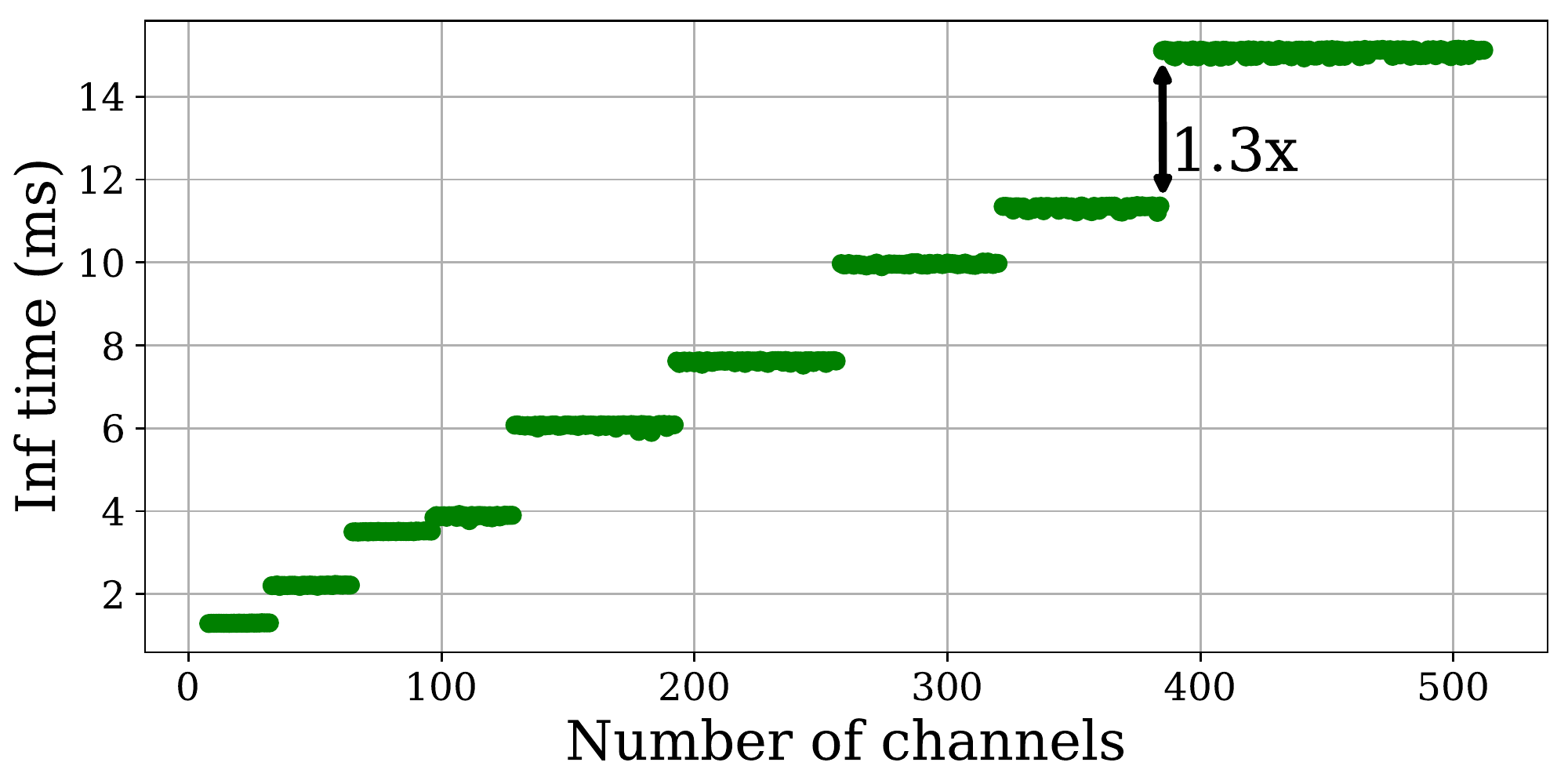}
    \caption{Execution pattern observed for channel pruning layer 14 of ResNet-50 implemented with CuDNN on \textit{Jetson Nano}.}
    \label{fig:nano_stairs}
\end{figure}

Figure~\ref{fig:nano_stairs} shows the staircase pattern of channel pruning on the Jetson Nano, also implemented with the CuDNN library. The same pattern exists on this device as observed in Figure~\ref{fig:tx2_layer14} for the Jetson TX2, due to similar GPU architectures, making performance modeling between the two easier. Other patterns are similar across all layers of ResNet as well as for VGG and AlexNet.

\begin{figure}
    \centering
    \includegraphics[clip,trim=11cm 0cm 0cm 0cm, height=1.8in]{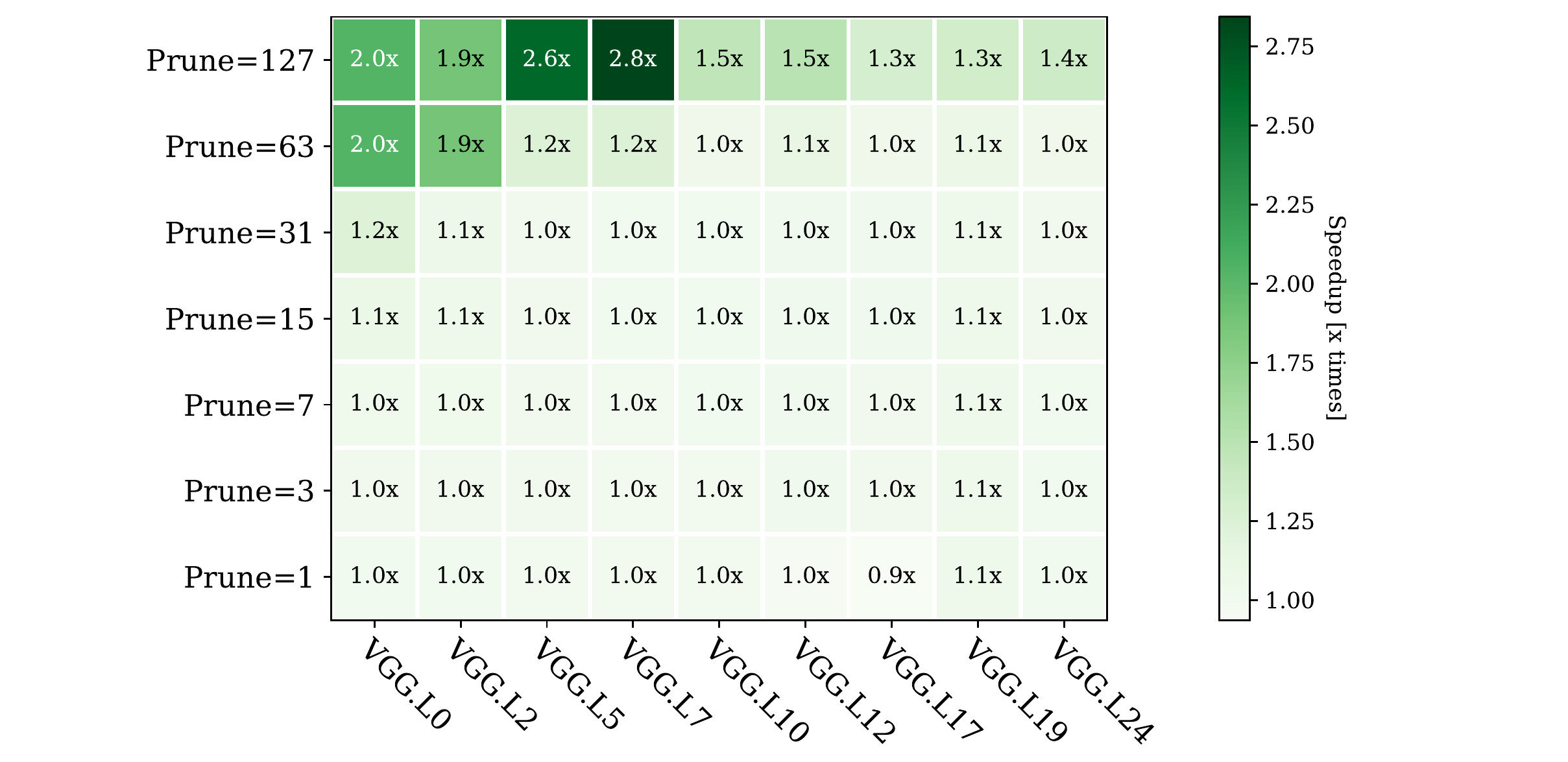}
    \caption{Speedups observed when pruning at different distances within each layer of  VGG-16 using the CuDNN implementation.}
    \label{fig:cudnn_vgg_heatmap}
\end{figure}

\begin{figure}
    \centering
    \includegraphics[clip,trim=15cm 0cm 0cm 0cm, height=1.8in]{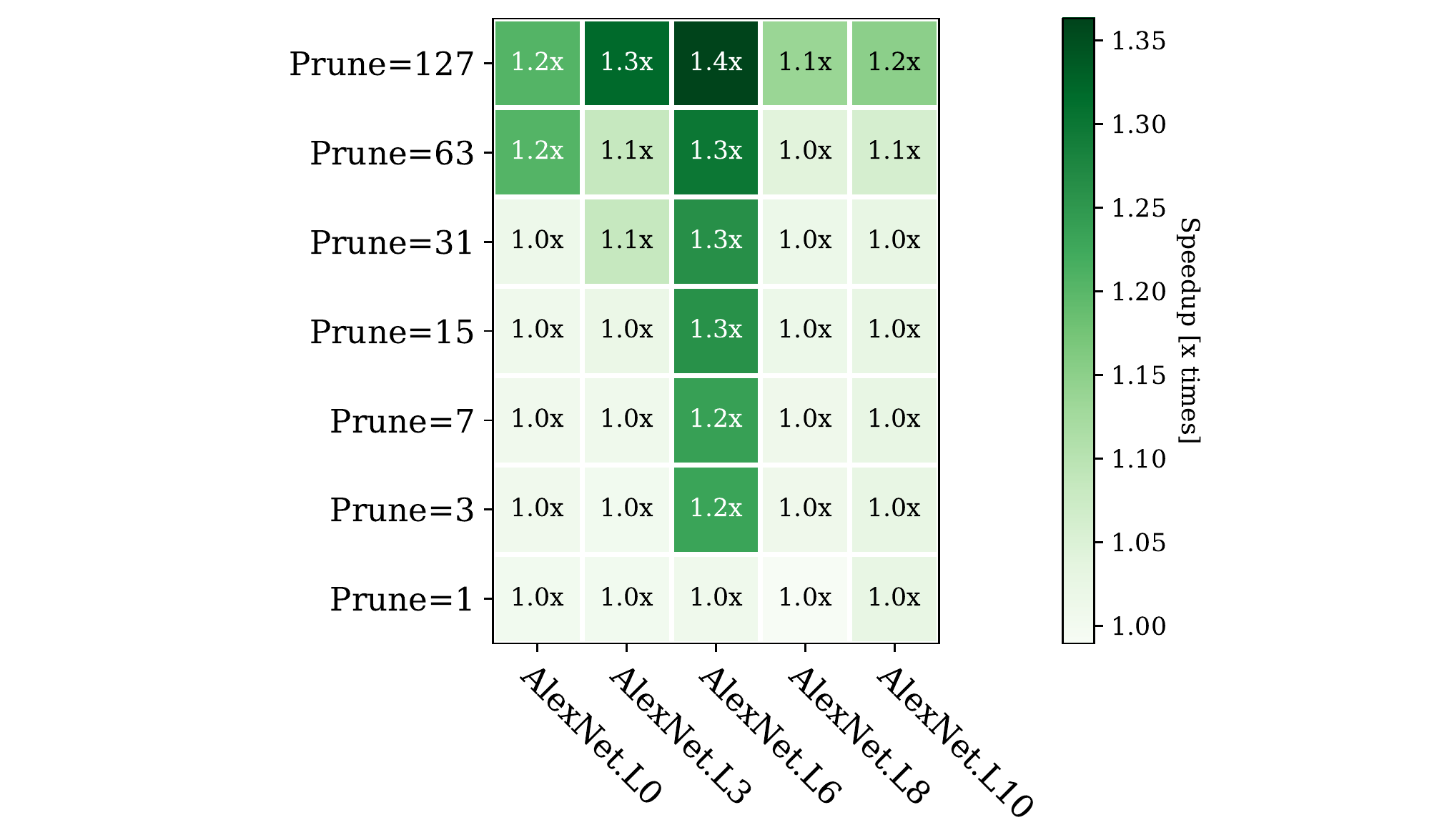}
    \caption{Speedups observed when pruning at different distances within each layer of  AlexNet using the CuDNN implementation.}
    \label{fig:cudnn_alex_heatmap}
\end{figure}

Different speedups can be achieved by different levels of pruning on each layer of a neural network, as presented in Figure~\ref{fig:cudnn_resnet_heatmap} for ResNet-50. This shows that some layers start experiencing speedups at a distance of 64 pruned channels and further, while for other it takes more pruning (due to layer input size and filter shape playing a role), and that speedups and gaps between stairs being uneven across layers. At a distance of 128 pruned channels, the maximum speedup for layer 16 is 3.3$\times$. A similar performance can be observed for the other two networks VGG-16 (Figure~\ref{fig:cudnn_vgg_heatmap}) and AlexNet (Figure~\ref{fig:cudnn_alex_heatmap}).

%%%%%%%%%%%%%%%%%%%%%%%%%%%%%%%%%%%%%%%%%

\subsubsection{Arm Compute Library using the Direct Convolution}

\begin{figure*}
    \centering
    \includegraphics[width=0.85\textwidth]{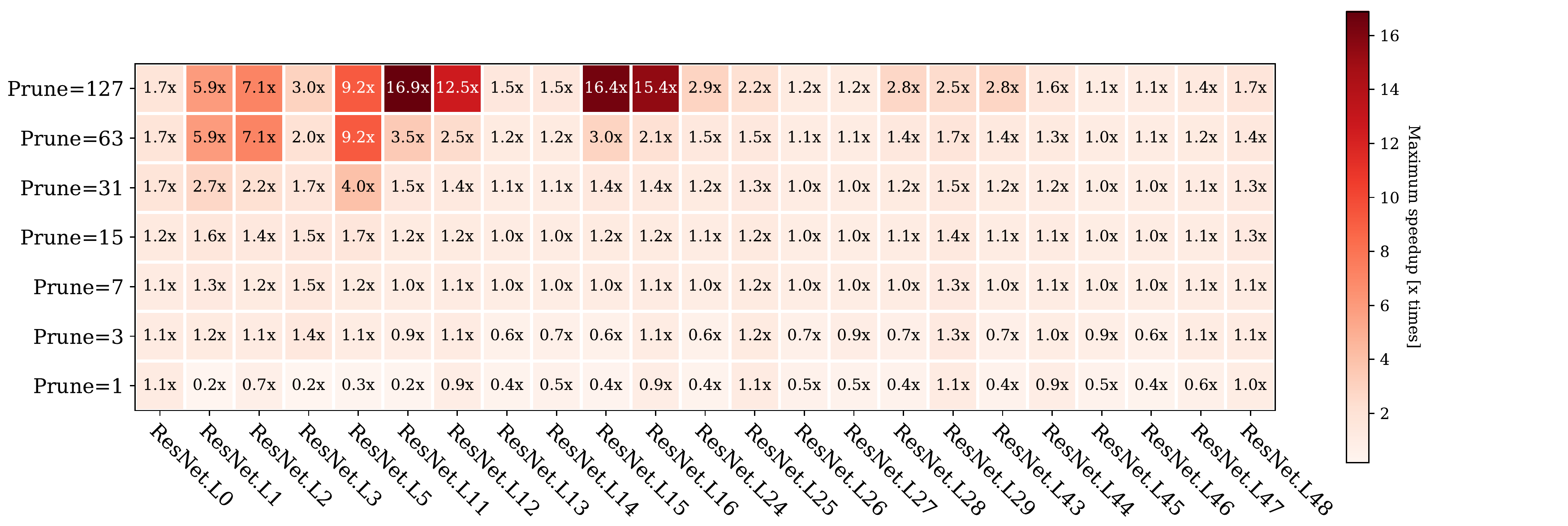}
    \caption{Speedups observed when pruning at different distances within each layer of  ResNet-50 using the Arm Compute Library Direct convolution implementation running on HiKey 970.}
    \label{fig:direct_heatmap}
\end{figure*}

\begin{figure}
    \centering
    \includegraphics[clip,trim=11cm 0cm 0cm 0cm, height=1.8in]{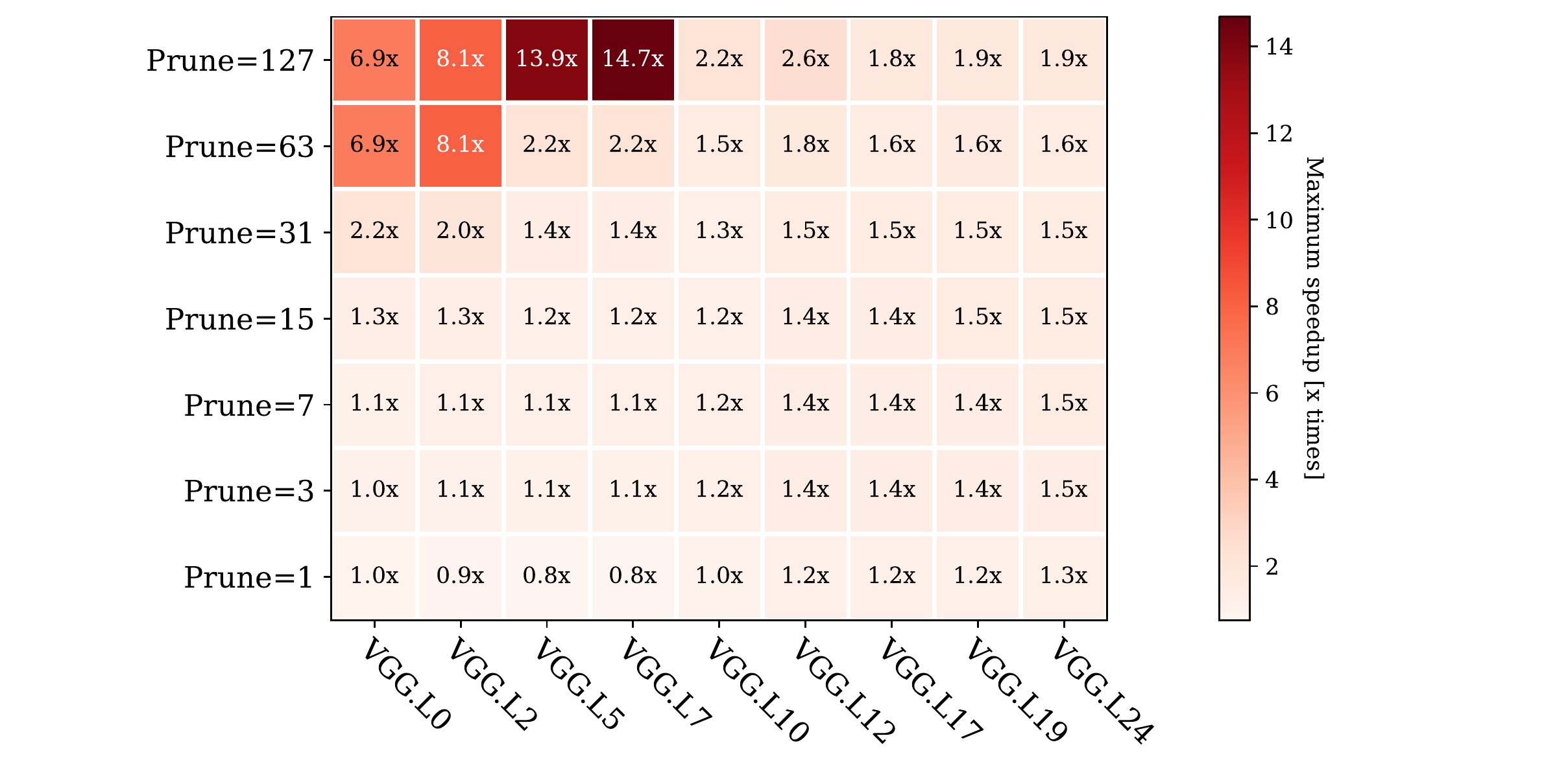}
    \caption{Speedups observed when pruning at different distances within each layer of  VGG-16 using the Arm Compute Library Direct convolution implementation.}
    \label{fig:heat_direct_vgg}
\end{figure}

In many cases where memory is tightly limited, Direct Convolution is the only option to implement a convolutional layer, due to GEMM expanding the matrix of input patches, which requires almost one order of magnitude more memory for a 3$\times$3 filter, as in the ResNet-50 and in other networks. Here we empirically explore the heuristics adopted in the ACL for these optimizations.

\begin{figure}
    \centering
    \includegraphics[width=0.9\columnwidth]{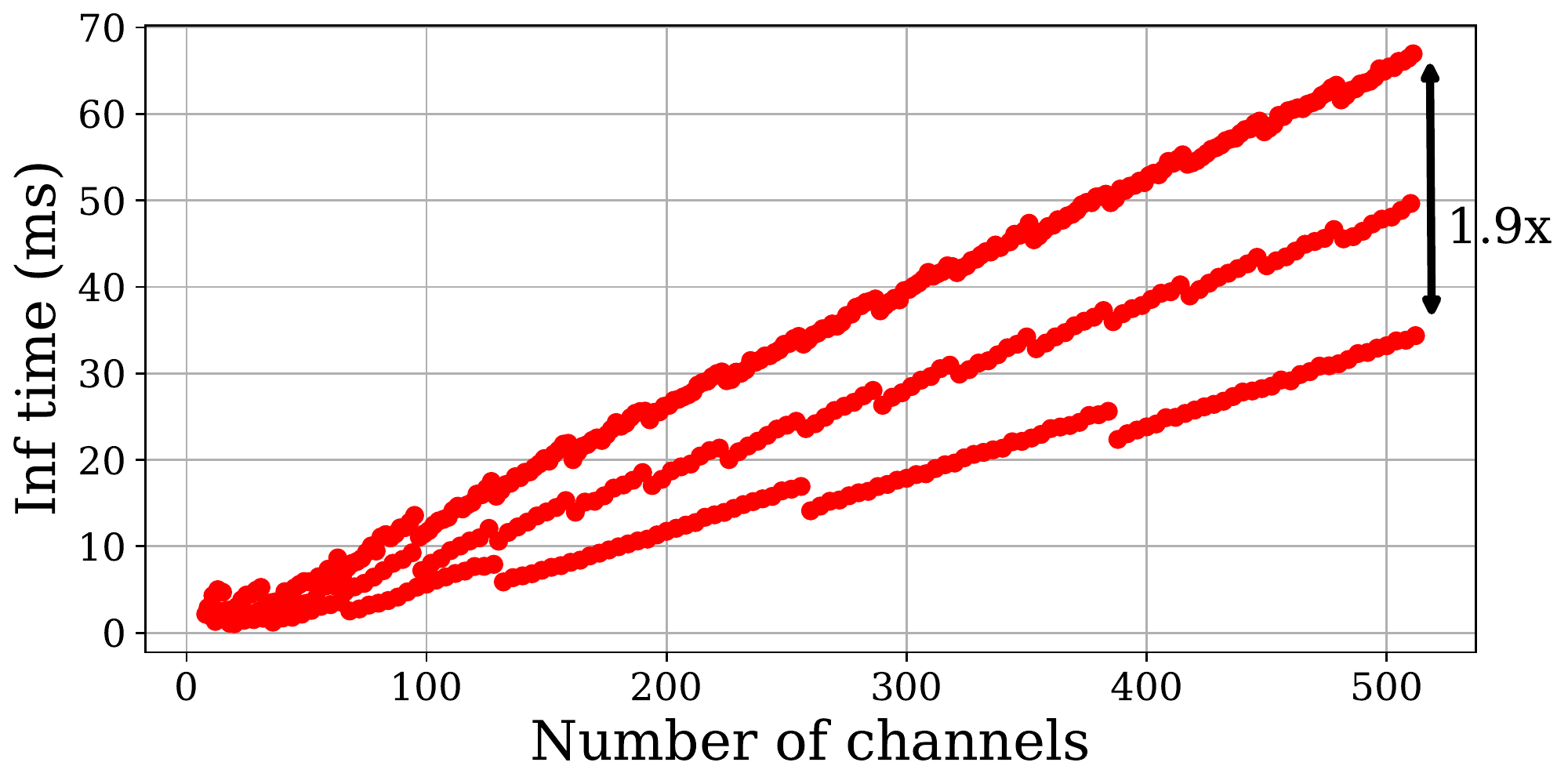}
    \caption{Execution pattern observed for channel pruning of ResNet-50 layer 14 implemented with Arm Compute Library Direct Convolution on HiKey 970 Mali GPU.}
    \label{fig:direct_stair}
\end{figure}

Figure~\ref{fig:direct_stair} shows that these heuristics lead to three execution levels alternating for different channel sizes of ResNet-50 layer 15. Having a linear pattern was expected, since each channel incrementally adds extra work in the deep nested loop of Direct convolution, however the three execution levels with up to 1.9$\times$ performance difference is unintuitive, and we explore this further in this section with the GPU simulator.

Pruning by just one channel for most of ResNet-50 layers shows a sub-unit speedup (or in actual terms a slowdown) as presented in Figure~\ref{fig:direct_heatmap}, going as low as 0.2$\times$ speedup or 80\% drop in performance, which is substantial. This indicates to us that optimization heuristics in the ACL are tuned for the standard shape of most popular neural networks, with even a small drop in the number of channels per layer leading to bad decisions from the built-in optimizer. A similar situation is observed for VGG-16 evaluated under the same conditions with the Direct Convolution of ACL (Figure~\ref{fig:heat_direct_vgg}). Similar patterns were observed when running both on the HiKey 970 and on the Odroid XU4. Considering that Direct Convolution is generally slower than all the other methods, it is understandable that not much development effort has been invested in optimizing this, although for many small devices with limited memory space this may be the only method that can actually execute at all.

%%%%%%%%%%%%%%%%%%%%%%%%%%%%%%%%%%%%%%%%%%%%%%%%%%%%%%%

\subsubsection{Arm Compute Library using the GEMM method}

\begin{figure*}
    \centering
    \includegraphics[width=0.85\textwidth]{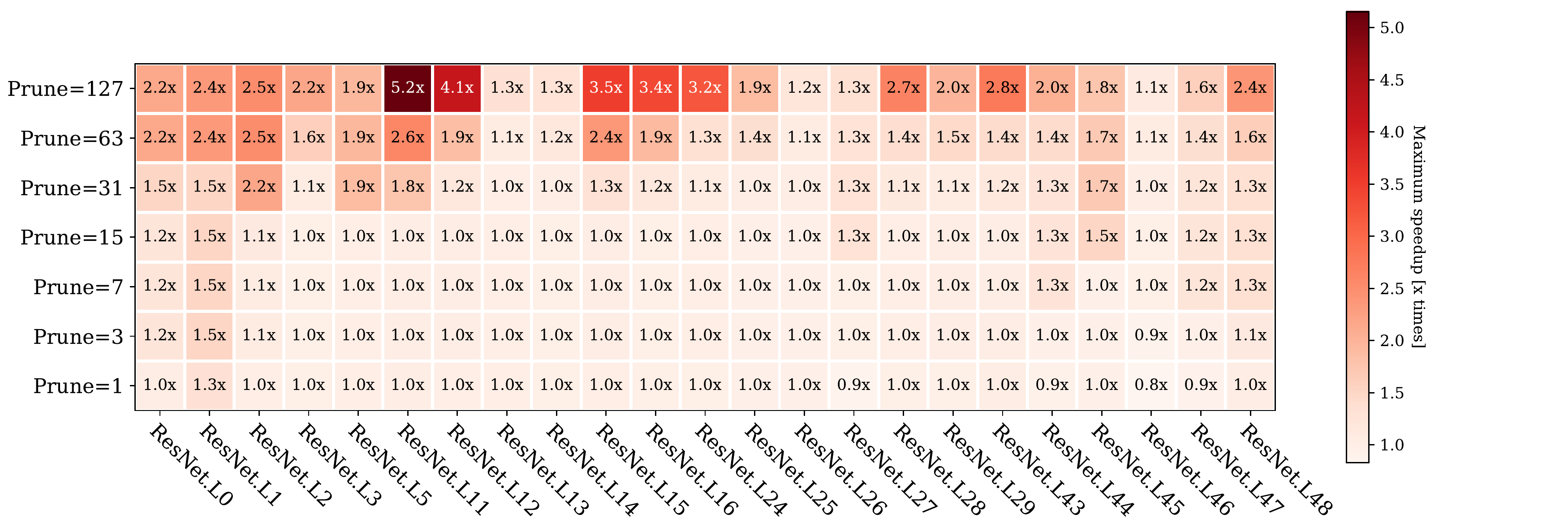}
    \caption{Speedups observed when pruning at different distances within each layer of  ResNet-50 using the Arm Compute Library GEMM implementation running on HiKey 970.}
    \label{fig:heat_gemm_resnet_hikey}
\end{figure*}

\begin{figure}
    \centering
    \includegraphics[width=0.9\columnwidth]{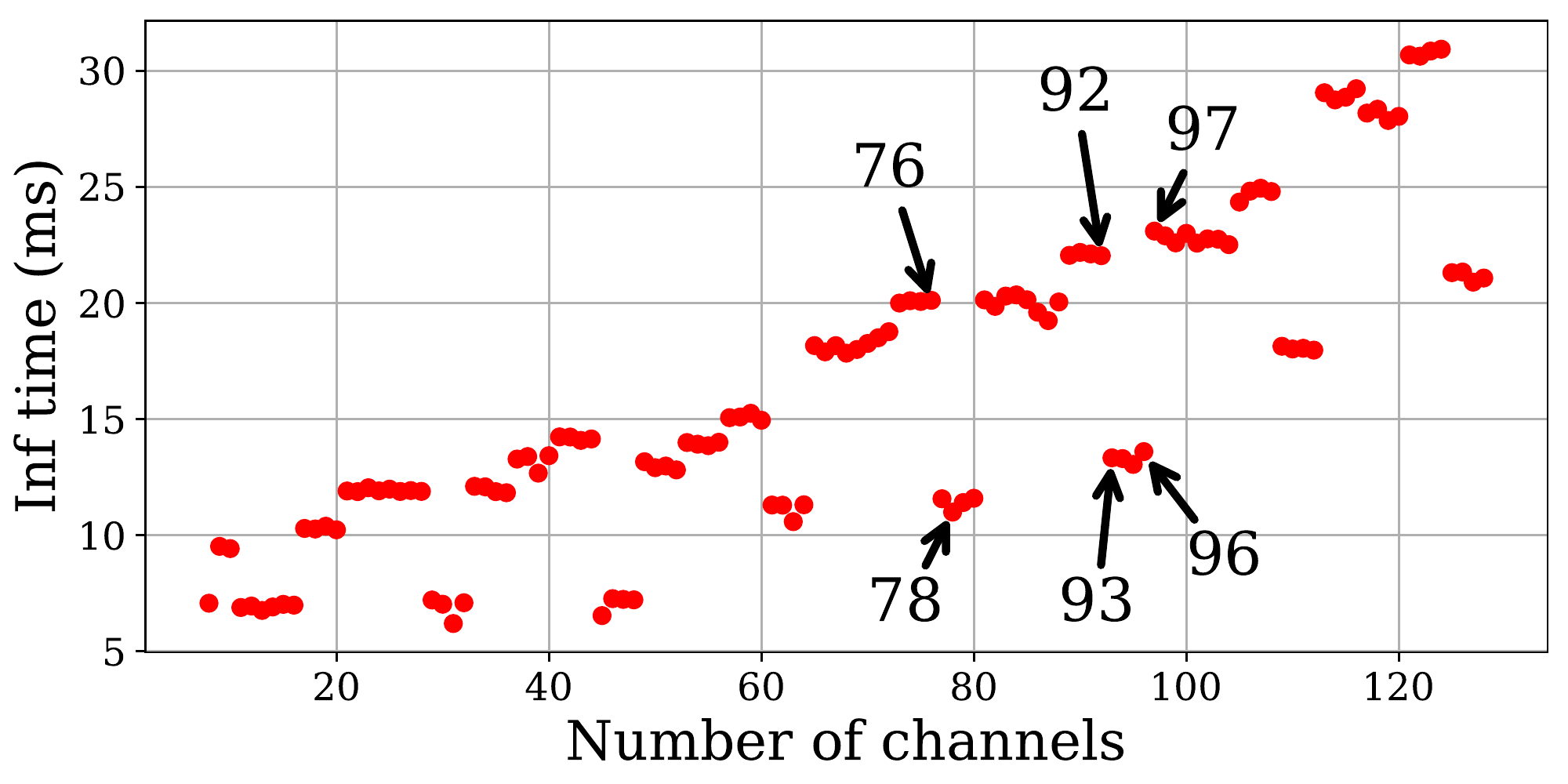}
    \caption{Execution pattern observed for channel pruning of ResNet-50 layer 16 implemented with Arm Compute Library GEMM on HiKey 970 Mali GPU.}
    \label{fig:gemm_annotations}
\end{figure}

\begin{figure}
    \centering
    \includegraphics[width=0.9\columnwidth]{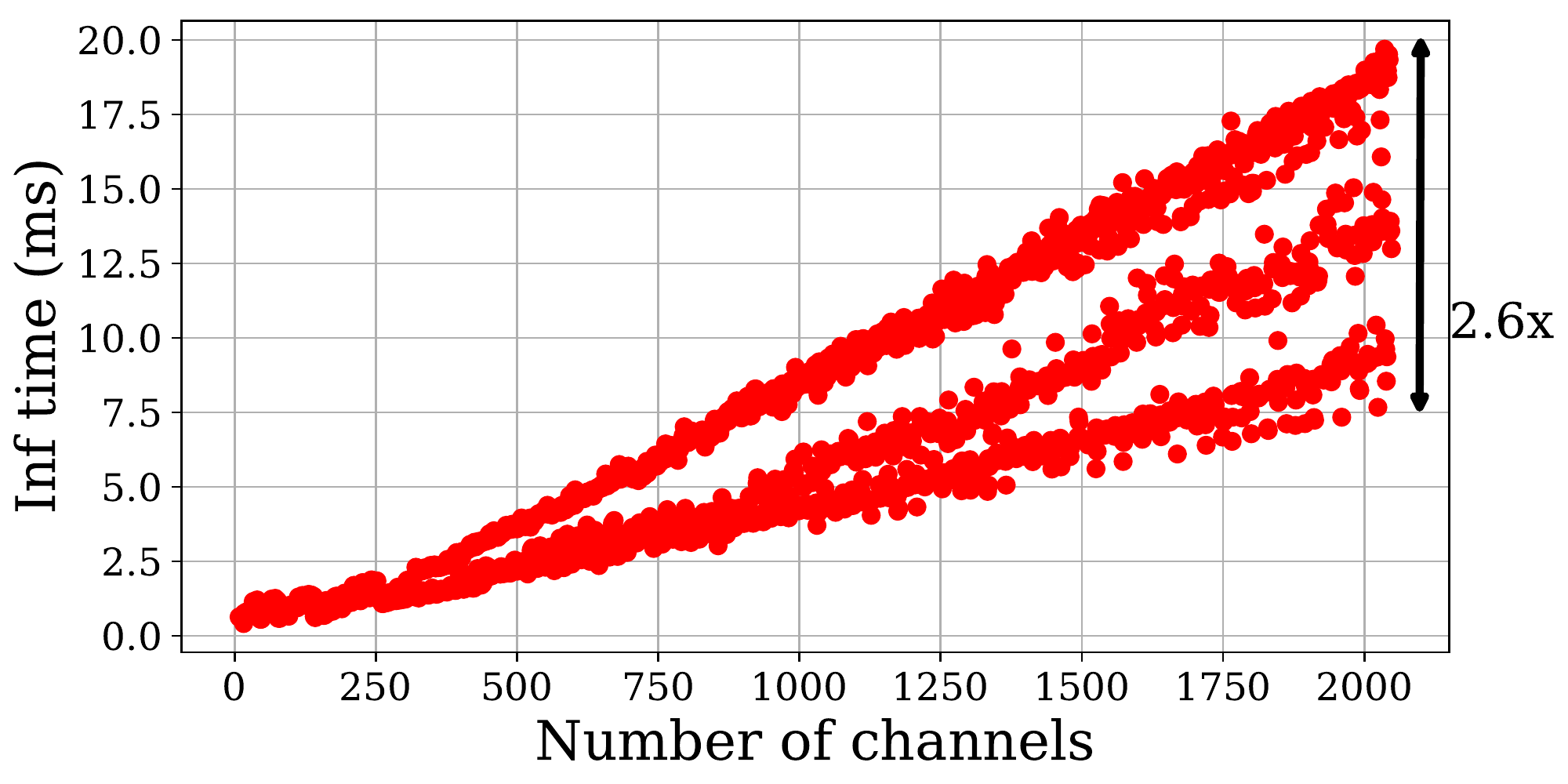}
    \caption{Large gap in inference time between small variations in the number of channels using the GEMM implementation with Arm Compute Library on layer 45 of ResNet-50.}
    \label{fig:g_l45_annotated}
\end{figure}

\begin{figure}
    \centering
    \includegraphics[clip,trim=11cm 0cm 0cm 0cm, height=1.8in]{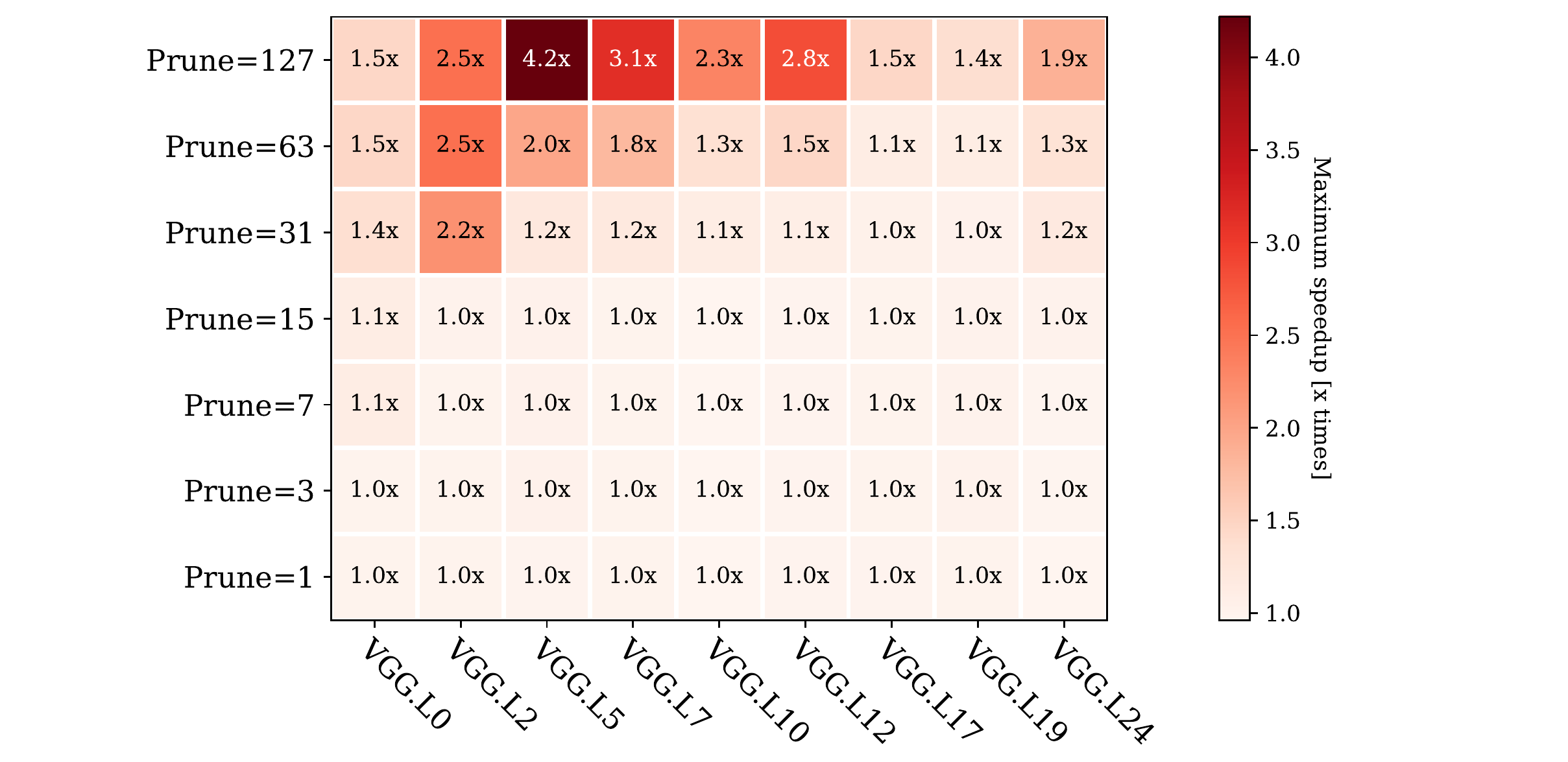}
    \caption{Speedups observed when pruning at different distances within each layer of  VGG-16 using the Arm Compute Library GEMM implementation.}
    \label{fig:heat_gemm_vgg}
\end{figure}

\begin{figure}
    \centering
    \includegraphics[clip,trim=15cm 0cm 0cm 0cm, height=1.8in]{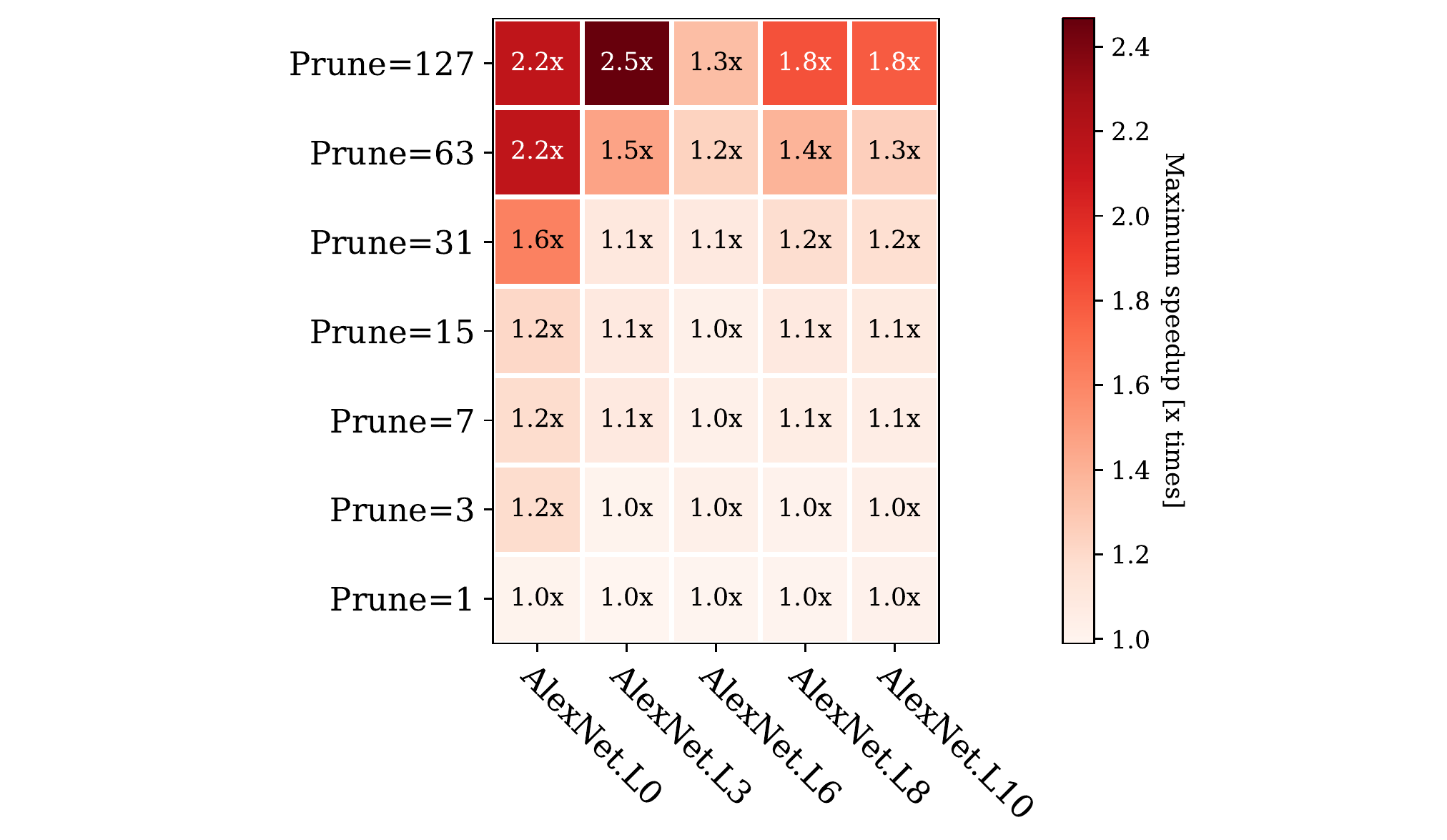}
    \caption{Speedups observed when pruning at different distances within each layer of  AlexNet using the Arm Compute Library GEMM implementation.}
    \label{fig:heat_gemm_alex}
\end{figure}

A more popular and faster approach for performing the convolutional workload is through GEMM which is also available in ACL. We run the pruned layers with a GEMM implementation, observing some unintuitive patterns.

Figure~\ref{fig:gemm_annotations} presents the execution time pattern for layer 16 of ResNet-50. Although we see similiar steps to those in the cuDNN implementation (which uses an optimised GEMM variant), this implementation of ACL presents two parallel staircases. Also observed from this is that each level is in groups of 4 which matches the size of vectorization, with channels 93 to 96 executing in 14 ms, while near channel sizes 92 and 97 jumping to 23 ms. Another observation is that between 76 and 78 channels (with only just 2 channels difference) inference time is improved from 20.12 ms to 10.996 ms, a 1.83$\times$ speedup between the two sizes.

An even wider gap in inference time between close number of channels is observed for layer 45, with 2036 channels inference is performed in 19.69 ms, while for 2024 channels this is performed in 7.67 ms, with a speedup of 2.57$\times$, as presented in Figure~\ref{fig:g_l45_annotated}.

Similarly to previous implementations, GEMM achieves a speedup of 5$\times$ for some layers of ResNet-50 for different levels of pruning (Figure~\ref{fig:heat_gemm_resnet_hikey}). Relevant to observe here is that there is no slowdown in the vicinity of the initial number of channels as observed for the Direct convolution, showing that heuristics for this optimization are uniformly modeled for different sizes. This is also observed for the other two networks VGG-16 (Figure~\ref{fig:heat_gemm_vgg}) and AlexNet (Figure~\ref{fig:heat_gemm_alex}).

%%%%%%%%%%%%%%%%%%%%%%%%%%%%%%

\subsubsection{TVM OpenCL Code Generator}

An atypical behavior pattern is observed with code generated by the TVM library. This shows a hybrid behavior between the Direct Convolution implementation of ACL and the GEMM implementation of ACL. Figure~\ref{fig:tvm_annotated} presents the execution time of pruned layer 14 from ResNet-50. While most channel counts are optimised with the GEMM implementation, there is a significant number of optimization calls instructed to use direct convolution which we know is generally slower, independent of the underlying hardware specifications. These occasional bad decisions are also observed on the other Mali platforms (Odroid XU4), leading to dramatic drops in performance, up to 13$\times$ as observed from Figure~\ref{fig:heat_tvm_resnet_hikey} for some layers. This may also be due to the version of the library, with dynamic developments happening in this space.

%%%%%%%%%%%%%%%%%%%

\input{gpu_simulation.tex}

%% file: gpu_simulation.tex
\begin{table}[h]
\centering
\caption{Arm Compute Library execution for layer 16 of ResNet-50 with 92 output channels.}
\small
\rowcolors{2}{white}{light-gray}
\begin{tabular}{lrr}
\thickhline
\bf{Kernel Name} & \bf{No Arithm. Instr.} & \bf{No Mem. Instr.}\\ \hline
im2col3x3\_nhwc & 1,365,198 & 212,152       \\ 
reshape\_to\_columns & 44,183,104 & 3,615,808       \\ 
gemm\_mm & 706,713,280 & 36,267,840       \\
gemm\_mm & 106,006,992 & 5,440,176       \\ \thickhline
\end{tabular}
\label{tab:92c}
\end{table}

\begin{table}[h]
\centering
\caption{Arm Compute Library execution for layer 16 of ResNet-50 with 93 output channels.}
\small
\rowcolors{2}{white}{light-gray}
\begin{tabular}{lrr}
\thickhline
\bf{Kernel Name} & \bf{No Arithm. Instr.} & \bf{No Mem. Instr.}\\ \hline
im2col3x3\_nhwc & 1,379,034 & 214,458       \\ 
reshape\_to\_columns & 44,183,104 & 3,615,808       \\ 
gemm\_mm & 848,055,936 & 43,521,408 \\ \thickhline
\end{tabular}
\label{tab:93c}
\end{table}

\begin{table}[h]
\centering
\caption{Arm Compute Library execution for layer 16 of ResNet-50 with 96 output channels.}
\small
\rowcolors{2}{white}{light-gray}
\begin{tabular}{lrr}
\thickhline
\bf{Kernel Name} & \bf{No Arithm. Instr.} & \bf{No Mem. Instr.}\\ \hline
im2col3x3\_nhwc & 1,420,542 & 221,376       \\ 
reshape\_to\_columns & 44,183,104 & 3,615,808       \\ 
gemm\_mm & 848,055,936 & 43,521,408       \\ \thickhline
\end{tabular}
\label{tab:timewindow_96}
\end{table}

\begin{table}[h]
\centering
\caption{Arm Compute Library execution for layer 16 of ResNet-50 with 97 output channels.}
\small
\rowcolors{2}{white}{light-gray}
\begin{tabular}{lrr}
\thickhline
\bf{Kernel Name} & \bf{No Arithm. Instr.} & \bf{No Mem. Instr.}\\ \hline
im2col3x3\_nhwc & 1,434,378 & 223,682       \\ 
reshape\_to\_columns & 44,183,104 & 3,615,808       \\ 
gemm\_mm & 848,055,936 & 43,521,408       \\
gemm\_mm & 35,335,664 & 1,813,392       \\ \thickhline
\end{tabular}
\label{tab:timewindow_97}
\end{table}

\subsection{Channel Pruning Observed Through GPU Simulation}\label{sec:simulation}

% \subsection{Introduction}
% With the increased convenience of higher level libraries, programmers are trading off direct control over the GPU. For example, when using the Arm Compute Library, there are unexplained performance differences as we vary the number of channels. 

Through the use of higher level libraries, like the ACL, we lose observability that we would normally have when working directly with OpenCL. To understand all the calls and kernel management performed by the Arm Compute Library for different sizes of a convolutional layer, as well as lower-level details about the execution in hardware, we executed the workloads in a Full-System Mali GPU simulator\cite{kaszyk2019full}.

As observed in previous experiments, there are unexplained performance differences when we vary the number of channels. In this section, we present our analysis of simulation results for GEMM and Direct Convolution implementations using the Mali GPU Simulator, and relate these directly to runtimes on the Hikey-970.

\subsubsection{Simulating the GEMM Method}

\begin{figure}
    \centering
    \includegraphics[width=\columnwidth]{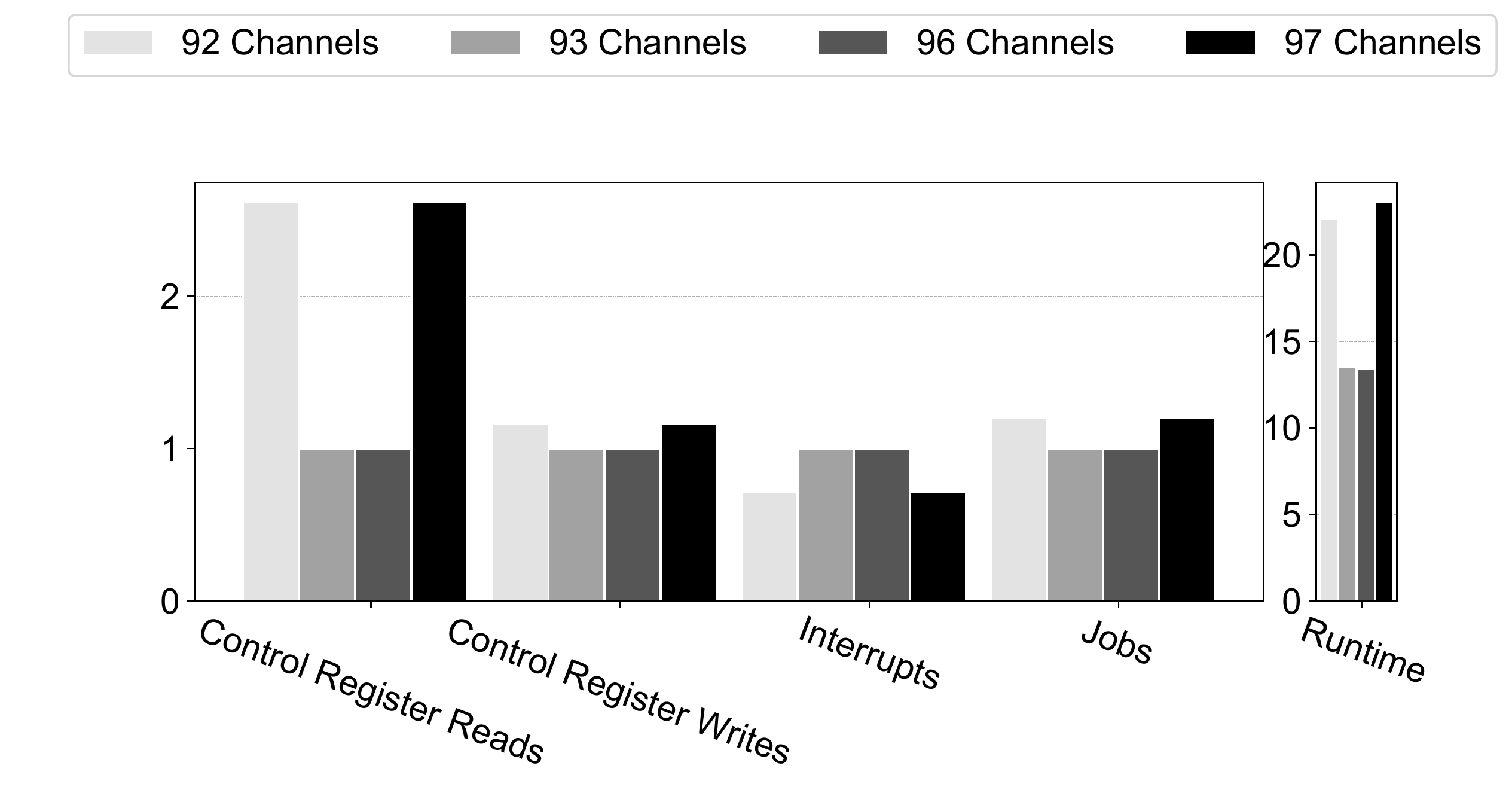}
    \caption{Relative System-Level Results for the GEMM implementation using 96 and 97 channels compared to runtimes on Hikey-970 board.}
    \label{fig:sys_level}
\end{figure}

The GEMM method is performed with the 32-bit Arm Compute Library Bifrost implementation. In hardware (Figure~\ref{fig:g_l45_annotated}), we observe that inference time dramatically drops when using 93 channels vs. 92 channels, and back up between layer configurations with 96 and with 97 channels. %, meaning that we get an improvement in accuracy at reduced computational cost. 
Using our OpenCL profiling tool, we can see that all dispatched kernels are the same between the two versions. Upon further inspection with our GPU simulator we can see that when using 93 channels, the number of jobs dispatched to the GPU is the same as the number of OpenCL calls made (OpenCL calls were observed with a profiling tool). However, when using 92 channels, additional jobs are dispatched to the GPU, meaning that the OpenCL runtime makes the decision to split the work. In Figure~\ref{fig:sys_level} we show the differences in number of jobs executed, as well as additional system-level results. Additional job creation and dispatch requires further communication between the CPU and GPU, and adds to the initialization cost on the GPU. This overhead often outweighs the benefits of dispatching workloads to accelerators. The difference in executed instructions is shown in Tables~\ref{tab:92c} and~\ref{tab:93c} for 92 and 93 channels and similarly for configurations with 96 and 97 channels in Tables~\ref{tab:timewindow_96} and \ref{tab:timewindow_97}.  While the im2col and reshape\_to\_columns kernels remain relatively steady while we vary the number of channels, the number of instructions in the gemm\_mm kernel increases by 4.35\%. The bulk of the computation for the gemm\_mm kernel however, is done in the first kernel, while the second kernel is responsible for only 13\% of the computation, showing the scope for improvement.

%%\begin{figure}
%    \centering
%    \includegraphics[width=\columnwidth]{scripts/local_global_ratios.pdf}
%    \caption{Correlations between local and global sizes and runtimes.}
%    \label{fig:local_global}
%\end{figure}

\subsubsection{Simulating the Direct Convolution Method}

\begin{table}[h!]
\centering
\caption{Arm Compute Library Direct Convolution work group sizes identified using GPU Simulator vs. Runtime measured on Hikey-970.}
\small
\rowcolors{2}{white}{light-gray}
\begin{tabular}{lrrrrr}
\thickhline
\bf{Number of} & \bf{X} & \bf{Y} & \bf{Z} & \bf{Relative Executed} & \bf{Time} \\
\bf{Channels} &  & &  & \bf{GPU Instructions} & \\ \hline
90 & 2 & 1 & 8 & 1.0 & 167.8716 \\ 
91 & 1 & 1 & 8 & 1.011 & 198.0468 \\
92 & 4 & 1 & 1 & 1.023 & 168.8311 \\
93 & 1 & 1 & 8 & 1.034 & 202.7299 \\

\end{tabular}
\label{tab:work_group_sizes}
\end{table}

In the direct convolution implementation, we no longer see differences in the number of jobs dispatched, however we still see differences in performance. OpenCL work group size selection is critical to performance, as it heavily impacts scheduling and caching on the GPU. \cite{petoumenosautotuning} shows that auto-tuning the OpenCL work group size provides mean speedup of 3.79x over the baseline configuration. In our experiments, the selection of the work group size for the dispatched OpenCL programs is left to the Arm Compute Library, and is completely invisible to the user. Examining channels 90-93, we see a wide range of reported runtimes, despite the fact that the number of executed instructions only increases by approximately one percent with each added channel. However, we observe very different work-splitting paradigms between successive layer sizes. As shown in Table \ref{tab:work_group_sizes}, the slower instances (91,93), use work group dimensions 1x1x8, while 90 and 92 channels use 2x1x8 and 4x1x1 respectively. Auto-tuning of the workloads and examining the effects of scheduling and caching have been left for future work.

%% file: discussion.tex
\section{Discussion} \label{sec:discussion}
%{\color{red}[Budget 0.75 pages.]}

\begin{figure*}
    \centering
    \includegraphics[width=0.85\textwidth]{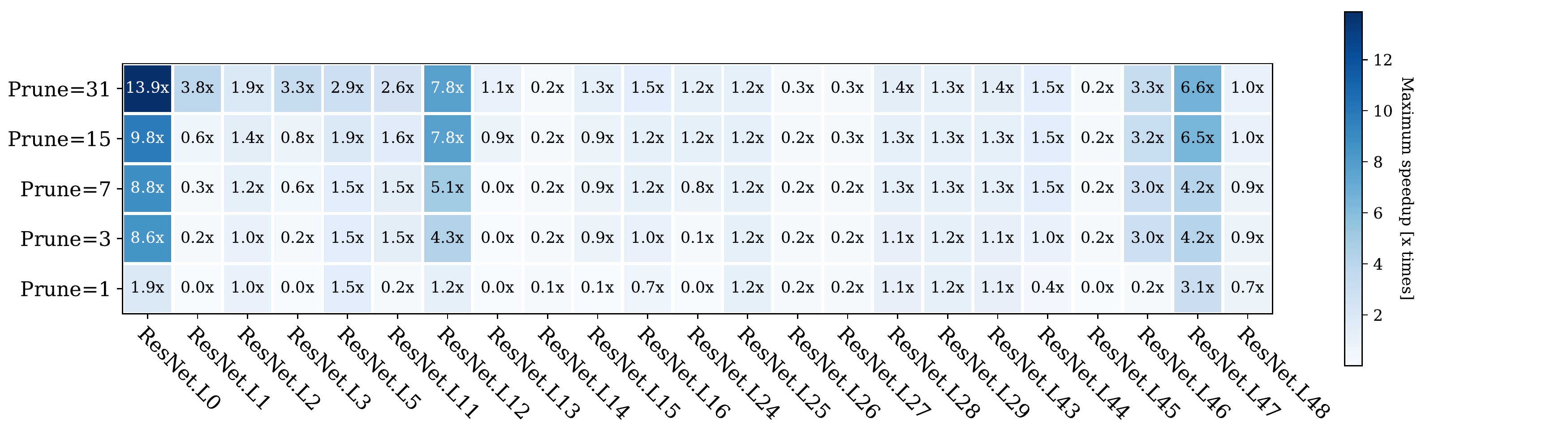}
    \caption{Speedups observed when pruning at different distances within each layer of  ResNet-50 using a TVM library implementation on HiKey 970.}
    \label{fig:heat_tvm_resnet_hikey}
    \vspace{1cm}
\end{figure*}
\begin{figure}
    \centering
    \includegraphics[width=0.9\columnwidth]{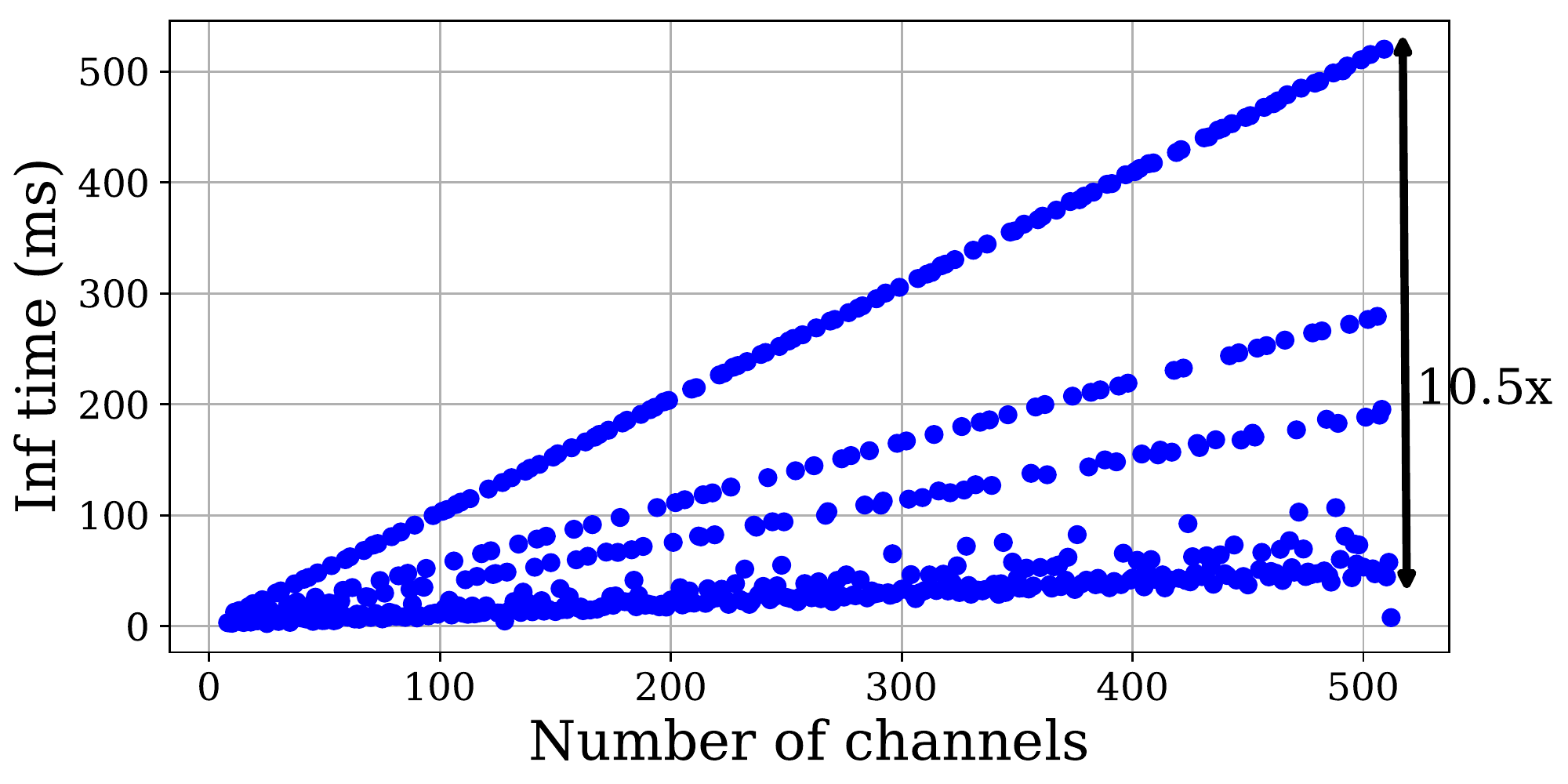}
    \caption{Layer 14 of ResNet-50 implemented with TVM OpenCL. Many sizes are untuned out of the box, showing a large variation due to uninstructed heuristics on HiKey 970.}
    \label{fig:tvm_annotated}
\end{figure}

Our exploration highlights some important limitations in deep learning libraries, showing that pre-designed heuristics fail for some arbitrary sizes of neural networks. As seen, pruning a number of channels can introduce slowdown rather than speedup, thus hurting performance, so these levels of pruning should be avoided. However, as expected, other pruning levels will run faster than the initial network configuration, where the library produces efficient GPU kernels. These optimal configurations can be found by profiling the kernel execution. These observations are relevant for pruning to the right number of channels and avoiding those levels that instruct optimizations which hurt performance. Instead by profiling, we can reduce the search space to the ones with superior speedup to test for accuracy in the network size-inference accuracy trade-off. Runtime optimal neural networks can be generated by coupling profiled performance on device with convolutional inference accuracy of pruned layers to instruct the best pruning level. We have initiated work in this direction showing that both execution time and inference accuracy can be considered simultaneously for efficient network compression to a target device~\cite{our18hakd}, although other research directions in library optimization and hardware design can also be considered.

From this exploration we find that no optimal library exists to outperform across all neural network layers. Neither Arm Compute Library, nor TVM dominates even with their auto-tuning enabled. Future solutions integrating optimizations from across different deep learning libraries could adapt their computation based on network and layer configuration to improve execution wit hardware aware performance. %For best performance, a combination of library selections at each layer offers the best performance, however efficient communication of data between these libraries should also be considered for further advances of efficient CNNs on mobile GPUs.

% {\color{red} TODO: \\
% - some observations about the underlying operations.\\
% - some observations about the library implementation/choices.
% }

%% file: related_work.tex
\section{Related Work}\label{sec:related_work}
%{\color{red}[Budget 0.5 pages.]}

%\subsection{Workloads on embedded GPUs}

Embedded GPUs are increasingly being considered for routine processing tasks. VComputeBench~\cite{8573477} is proposed as a set of benchmarks to help developers understand the differences in performance and portability of Vulkan, a new programming model for cross-platform GPGPU computing notably on mobile and embedded GPUs. The main conclusions are that performance improvements imply a high programming effort and performance portability for mobile GPU architectures is not guaranteed. Lee et al.~\cite{8247279} propose an aging-aware workload management technique for embedded GPUs in the presence of process variation. The simulation results show that the proposed technique improves the GPU aging over 95 percent of cases whereas the state-of-the-art compiler-based technique improves the GPU aging in 72.25 percent of cases. In~\cite{Otterness2016GPUSF} the authors analyze the interference in concurrent GPU computations for several image processing on NVIDIA Jetson TK1 and TX1 boards. The results suggest that allowing multiple kernels to be co-scheduled may have a positive impact on real-time schedulability. It would be good to observe the results on newer NVIDIA Jetson models like Tx2 and Xavier. Dev et al. \cite{7581285} analyze CPU-GPU processors and characterize OpenCL workloads during run-time with the target of mapping them to the appropriate device under time-varying physical (i.e., chip power limit) and CPU load conditions. The proposed scheduler provides average improvements of 31\% and 4\% in runtime and energy, respectively with respect to the state-of-the-art. However, the number of benchmarks analyzed in this work is very limited. In \cite{8047282} ButterFly is proposed as a novel system to collaboratively utilize mobile GPUs in order to process high-quality rendering details for on-the-go mobile users. Butterfly outperforms the performance of previous state-of-the-art systems and achieves more than 28.3\% power saving. The scalability of the proposed system is not completely clear, as only up to 3 collaborative workers are considered. In~\cite{multiprog_2019} the authors outline two methods for fast convolution on embedded GPUs, an iterative vectorized approach and a Morton GEMM based approach, providing 3x faster inference times than some current state-of-the-art systems. However, newer libraries like the Arm Compute Library are not considered in the study.

%\subsection{Profiling DNNs on GPU/CPU}

In~\cite{Mojumder2018ProfilingDW}, the authors profile and analyze the training process of five popular DNNs (GoogLeNet, AlexNet, Inception-v3, ResNet and LeNet) using 1, 2, 4 and 8 Volta-based GPUs. However, there is no discussion in the text about potential scalability effects beyond 8 GPUs. An extensive performance analysis and profiling of DNN training is performed in ~\cite{8573476}, where eight state-of-the-art DNN models are implemented on three major deep learning frameworks (TensorFlow, MXNet, and CNTK). The objective is to evaluate the efficiency of training for different hardware configurations (single-/multi-GPU and multi-machine). The scalability study of this work in terms of GPUs/machines is very limited (up to 4 GPUs/machines). The DeftNN~\cite{8686533} execution framework was used to profile 6 state-of-the-art DNNs running on an Nvidia Titan X (Pascal) GPU in order to automatically and transparently improve their execution performance. It would also be good to observe how the proposed technique translates to embedded GPUs for the for inference procedure. Lew et al.~\cite{8695671} modify GPGPU-Sim in order to study ML workloads and analyze their behavior (e.g. run applications that use cuDNN and PyTorch). The simulator provides detailed information about memory usage, power efficiency, etc. thus identifying opportunities for architectural optimization. The weak point of this work is that the number of workloads analyzed is very limited, which makes it difficult to extrapolate the conclusions. In~\cite{Awan:2017:IPC:3146347.3146356} the authors characterize the performance of DNN training for AlexNet and ResNet-50 for a wide-range of CPU and GPU architectures including the Intel Xeon Phi (Knights Landing) processors and NVIDIA Pascal GPUs. Given the range of platforms analyzed (CPUs and GPUs), it would be good to complement the study analyzing some more networks (currently only 2).

NUMA-Caffe~\cite{Roy:2018:NND:3212710.3199605} is proposed as a novel NUMA-aware framework for training CNNs on modern CPU-based multi- and many-core-based architectures. The authors apply system-level profiling tools to quantify the bottlenecking effects so as to explore the potential root cause of Caffe's scalability issue on multi-core systems. It would be interesting to observe how these scalability issues translate to other widely adopted frameworks like TensorFlow or Pytorch. In~\cite{8192500} the authors propose Scalpel to customize DNN pruning to the underlying hardware by matching the pruned network structure to the data-parallel hardware organization. Callgrind is used for profiling and determining the parallelism level of the target hardware platform. A weak point of this work is that most of the analysis provided is about small network models and datasets.

%% file: conclusions.tex
\section{Conclusions}\label{sec:conclusions}

Mobile and embedded GPUs are increasingly being used for running neural network workloads, through popular higher level libraries. However, these libraries exhibit behaviors that are not well understood, springing from their built-in heuristics for workload optimization. We show that channel pruning (a popular neural network compression technique to make models available on small devices) can lead to major slowdowns, up to 2$\times$, when performed without considering the library and device performance. We expose the behavior of three popular deep learning libraries for embedded GPUs, Arm Compute Library, CuDNN and TVM, showing that a supervised selection of the number of channels (filters) can lead to speedups of 3$\times$ using CuDNN, and more than 10$\times$ using Arm Compute Library and TVM. %Variations in performance across libraries can be exploited by selecting the best implementations for each neural network layer of a neural network to optimize size and performance.
We believe these observations offer a direction for library developers to optimize the performance of their libraries for any shape of convolutional layers and for machine learning specialists to design neural networks that exploit the sweet-spots of library-hardware performance.

%% file: main.bbl
\begin{thebibliography}{10}

\bibitem{alexnet}
Sergey Ioffe and Christian Szegedy.
\newblock Imagenet classification with deep convolutional neural networks.
\newblock In {\em International Conference on Machine Learning}, pages
  448--456, 2015.

\bibitem{He_2016_CVPR}
Kaiming He, Xiangyu Zhang, Shaoqing Ren, and Jian Sun.
\newblock Deep residual learning for image recognition.
\newblock In {\em The IEEE Conference on Computer Vision and Pattern
  Recognition (CVPR)}, June 2016.

\bibitem{denseNet2017}
G.~Huang, Z.~Liu, L.~v.~d. Maaten, and K.~Q. Weinberger.
\newblock Densely connected convolutional networks.
\newblock In {\em 2017 IEEE Conference on Computer Vision and Pattern
  Recognition (CVPR)}, pages 2261--2269, July 2017.

\bibitem{7298965}
J.~Long, E.~Shelhamer, and T.~Darrell.
\newblock Fully convolutional networks for semantic segmentation.
\newblock In {\em 2015 IEEE Conference on Computer Vision and Pattern
  Recognition (CVPR)}, pages 3431--3440, June 2015.

\bibitem{10.1007/978-3-319-24574-4_28}
Olaf Ronneberger, Philipp Fischer, and Thomas Brox.
\newblock U-net: Convolutional networks for biomedical image segmentation.
\newblock In Nassir Navab, Joachim Hornegger, William~M. Wells, and
  Alejandro~F. Frangi, editors, {\em Medical Image Computing and
  Computer-Assisted Intervention -- MICCAI 2015}, pages 234--241, Cham, 2015.
  Springer International Publishing.

\bibitem{7485869}
S.~Ren, K.~He, R.~Girshick, and J.~Sun.
\newblock Faster r-cnn: Towards real-time object detection with region proposal
  networks.
\newblock {\em IEEE Transactions on Pattern Analysis and Machine Intelligence},
  39(6):1137--1149, June 2017.

\bibitem{yolo2016}
Joseph Redmon, Santosh Divvala, Ross Girshick, and Ali Farhadi.
\newblock Deep residual learning for image recognition.
\newblock In {\em The IEEE Conference on Computer Vision and Pattern
  Recognition (CVPR)}, 2016.

\bibitem{7780634}
L.~A. Gatys, A.~S. Ecker, and M.~Bethge.
\newblock Image style transfer using convolutional neural networks.
\newblock In {\em 2016 IEEE Conference on Computer Vision and Pattern
  Recognition (CVPR)}, pages 2414--2423, June 2016.

\bibitem{Zhang2016TowardsES}
Ying Zhang, Mohammad Pezeshki, Philemon Brakel, Saizheng Zhang, C{\'e}sar
  Laurent, Yoshua Bengio, and Aaron~C. Courville.
\newblock Towards end-to-end speech recognition with deep convolutional neural
  networks.
\newblock In {\em INTERSPEECH}, 2016.

\bibitem{cnnsentence2014}
Yoon Kim.
\newblock Convolutional neural networks for sentence classification.
\newblock In {\em The Conference on Empirical Methods in Natural Language
  Processing}, 2014.

\bibitem{Kalchbrenner14aconvolutional}
Nal Kalchbrenner, Edward Grefenstette, and Phil Blunsom.
\newblock A convolutional neural network for modelling sentences.
\newblock In {\em In Proceedings of the 52nd Annual Meeting of the Association
  for Computational Linguistics}, 2014.

\bibitem{NIPS1988_156}
Stephen~Jose Hanson and Lorien~Y. Pratt.
\newblock Comparing biases for minimal network construction with
  back-propagation.
\newblock In D.~S. Touretzky, editor, {\em Advances in Neural Information
  Processing Systems 1}, pages 177--185. Morgan-Kaufmann, 1989.

\bibitem{han2015deep}
Song Han, Huizi Mao, and William~J Dally.
\newblock Deep compression: Compressing deep neural networks with pruning,
  trained quantization and huffman coding.
\newblock {\em arXiv preprint arXiv:1510.00149}, 2015.

\bibitem{han2015learning}
Song Han, Jeff Pool, John Tran, and William Dally.
\newblock Learning both weights and connections for efficient neural network.
\newblock In {\em Advances in Neural Information Processing Systems}, pages
  1135--1143, 2015.

\bibitem{he2017channel}
Yihui He, Xiangyu Zhang, and Jian Sun.
\newblock Channel pruning for accelerating very deep neural networks.
\newblock In {\em International Conference on Computer Vision (ICCV)},
  volume~2, page~6, 2017.

\bibitem{our18iiswc}
Jack Turner, Jos\'e Cano, Valentin Radu, Elliot~J Crowley, Michael O’Boyle,
  and Amos Storkey.
\newblock Characterising across-stack optimisations for deep convolutional
  neural networks.
\newblock In {\em Proc IISWC}. IEEE, 2018.

\bibitem{hu2018squeeze}
Jie Hu, Li~Shen, and Gang Sun.
\newblock Squeeze-and-excitation networks.
\newblock In {\em Proceedings of the IEEE conference on computer vision and
  pattern recognition}, pages 7132--7141, 2018.

\bibitem{jia2014learning}
Yangqing Jia.
\newblock {\em Learning semantic image representations at a large scale}.
\newblock PhD thesis, UC Berkeley, 2014.

\bibitem{our18hakd}
Jack Turner, Elliot~J Crowley, Valentin Radu, Jos\'e Cano, Amos Storkey, and
  Michael O'Boyle.
\newblock Distilling with performance enhanced students.
\newblock {\em CoRR}, 2018.

\bibitem{he2016deep}
Kaiming He, Xiangyu Zhang, Shaoqing Ren, and Jian Sun.
\newblock Deep residual learning for image recognition.
\newblock In {\em Proceedings of the IEEE conference on computer vision and
  pattern recognition}, pages 770--778, 2016.

\bibitem{vgg}
Karen Simonyan and Andrew Zisserman.
\newblock Very deep convolutional networks for large-scale image recognition.
\newblock In {\em International Conference on Learning Representations}, 2015.

\bibitem{kaszyk2019full}
Kuba Kaszyk, Harry Wagstaff, Tom Spink, Bj{\"o}rn Franke, Mike O'Boyle, Bruno
  Bodin, and Henrik Uhrenholt.
\newblock Full-system simulation of mobile cpu/gpu platforms.
\newblock In {\em 2019 IEEE International Symposium on Performance Analysis of
  Systems and Software (ISPASS)}, pages 68--78. IEEE, 2019.

\bibitem{petoumenosautotuning}
Chris Cummins~Pavlos Petoumenos, Michel Steuwer, and Hugh Leather.
\newblock Autotuning opencl workgroup size for stencil patterns.

\bibitem{8573477}
N.~{Mammeri} and B.~{Juurlink}.
\newblock Vcomputebench: A vulkan benchmark suite for gpgpu on mobile and
  embedded gpus.
\newblock In {\em 2018 IEEE International Symposium on Workload
  Characterization (IISWC)}, pages 25--35, Sep. 2018.

\bibitem{8247279}
H.~{Lee}, M.~{Shafique}, and M.~A. {Al Faruque}.
\newblock Aging-aware workload management on embedded gpu under process
  variation.
\newblock {\em IEEE Transactions on Computers}, 67(7):920--933, July 2018.

\bibitem{Otterness2016GPUSF}
Nathan Otterness, Vance Miller, Ming Yang, James~H. Anderson, Frank~Donelson
  Smith, and Shige Wang.
\newblock Gpu sharing for image processing in embedded real-time systems.
\newblock In {\em 12thAnnual Workshop onOperating Systems Platforms forEmbedded
  Real-Time Applications}, 2016.

\bibitem{7581285}
K.~{Dev}, X.~{Zhan}, and S.~{Reda}.
\newblock Power-aware characterization and mapping of workloads on cpu-gpu
  processors.
\newblock In {\em 2016 IEEE International Symposium on Workload
  Characterization (IISWC)}, pages 1--2, Sep. 2016.

\bibitem{8047282}
C.~{Wu}, B.~{Yang}, W.~{Zhu}, and Y.~{Zhang}.
\newblock Toward high mobile gpu performance through collaborative workload
  offloading.
\newblock {\em IEEE Transactions on Parallel and Distributed Systems},
  29(2):435--449, Feb 2018.

\bibitem{multiprog_2019}
Simon Rovder, Jos\'e Cano, and Michael O'Boyle.
\newblock {Optimising Convolutional Neural Networks Inference on Low-Powered
  GPUs}.
\newblock In {\em 12th International Workshop on Programmability and
  Architectures for Heterogeneous Multicores (MULTIPROG)}, January 2019.

\bibitem{Mojumder2018ProfilingDW}
Saiful~A. Mojumder, Marcia~S. Louis, Yifan Sun, Amir~Kavyan Ziabari,
  Jos{\'e}~L. Abell{\'a}n, John Kim, David~R. Kaeli, and Ajay~Jayant Joshi.
\newblock Profiling dnn workloads on a volta-based dgx-1 system.
\newblock {\em 2018 IEEE International Symposium on Workload Characterization
  (IISWC)}, pages 122--133, 2018.

\bibitem{8573476}
H.~{Zhu}, M.~{Akrout}, B.~{Zheng}, A.~{Pelegris}, A.~{Jayarajan},
  A.~{Phanishayee}, B.~{Schroeder}, and G.~{Pekhimenko}.
\newblock Benchmarking and analyzing deep neural network training.
\newblock In {\em 2018 IEEE International Symposium on Workload
  Characterization (IISWC)}, pages 88--100, Sep. 2018.

\bibitem{8686533}
P.~{Hill}, A.~{Jain}, M.~{Hill}, B.~{Zamirai}, C.~{Hsu}, M.~A. {Laurenzano},
  S.~{Mahlke}, L.~{Tang}, and J.~{Mars}.
\newblock Deftnn: Addressing bottlenecks for dnn execution on gpus via synapse
  vector elimination and near-compute data fission.
\newblock In {\em 2017 50th Annual IEEE/ACM International Symposium on
  Microarchitecture (MICRO)}, pages 786--799, Oct 2017.

\bibitem{8695671}
J.~{Lew}, D.~A. {Shah}, S.~{Pati}, S.~{Cattell}, M.~{Zhang}, A.~{Sandhupatla},
  C.~{Ng}, N.~{Goli}, M.~D. {Sinclair}, T.~G. {Rogers}, and T.~M. {Aamodt}.
\newblock Analyzing machine learning workloads using a detailed gpu simulator.
\newblock In {\em 2019 IEEE International Symposium on Performance Analysis of
  Systems and Software (ISPASS)}, pages 151--152, March 2019.

\bibitem{Awan:2017:IPC:3146347.3146356}
Ammar~Ahmad Awan, Hari Subramoni, and Dhabaleswar~K. Panda.
\newblock An in-depth performance characterization of cpu- and gpu-based dnn
  training on modern architectures.
\newblock In {\em Proceedings of the Machine Learning on HPC Environments},
  MLHPC'17, pages 8:1--8:8, New York, NY, USA, 2017. ACM.

\bibitem{Roy:2018:NND:3212710.3199605}
Probir Roy, Shuaiwen~Leon Song, Sriram Krishnamoorthy, Abhinav Vishnu, Dipanjan
  Sengupta, and Xu~Liu.
\newblock Numa-caffe: Numa-aware deep learning neural networks.
\newblock {\em ACM Trans. Archit. Code Optim.}, 15(2):24:1--24:26, June 2018.

\bibitem{8192500}
J.~{Yu}, A.~{Lukefahr}, D.~{Palframan}, G.~{Dasika}, R.~{Das}, and S.~{Mahlke}.
\newblock Scalpel: Customizing dnn pruning to the underlying hardware
  parallelism.
\newblock In {\em 2017 ACM/IEEE 44th Annual International Symposium on Computer
  Architecture (ISCA)}, pages 548--560, June 2017.

\end{thebibliography}
